\PassOptionsToPackage{table}{xcolor}
\documentclass[10pt,twocolumn,letterpaper]{article}

\usepackage[pagenumbers]{iccv} % To force page numbers, e.g. for an arXiv version
\usepackage{multirow}
\usepackage[table]{xcolor}
\usepackage{subcaption}
\usepackage{tagging}
\usepackage{enumitem}% http://ctan.org/pkg/enumitem
\usepackage{makecell} % Include this in the preamble
\usepackage[export]{adjustbox} % For precise alignment
\usepackage{pifont}% http://ctan.org/pkg/pifont
\usepackage{mathtools} % Add to preamble if not already present
\usepackage{marvosym}
\usepackage[accsupp]{axessibility}

\definecolor{iccvblue}{rgb}{0.21,0.49,0.74}
\usepackage[pagebackref,breaklinks,colorlinks,allcolors=iccvblue]{hyperref}

\def\confName{ICCV}
\def\confYear{2025}

\makeatletter
\renewcommand{\@fnsymbol}[1]{\ifcase#1\or \Letter\or *\or *\or \S\or \P\or \#\else @\fi}
\makeatother

\title{Rethinking Few Shot CLIP Benchmarks: A Critical Analysis in the \\ Inductive Setting}

\author{
Alexey Kravets, \quad 
Da Chen\thanks{Corresponding author}, \quad 
Vinay P. Namboodiri \\
University of Bath, UK \\
{\tt\small ak3095@bath.ac.uk}, 
{\tt\small da.chen@bath.edu}, 
{\tt\small vpn22@bath.ac.uk} 
}

\begin{document}

\maketitle

\begin{abstract}

CLIP is a foundational model with transferable classification performance in the few-shot setting. Several methods have shown improved performance of CLIP using few-shot examples. However, so far, all these techniques have been benchmarked using standard few-shot datasets. We argue that this mode of evaluation does not provide a true indication of the \textit{inductive} generalization ability using few-shot examples. As most datasets have been seen by the CLIP model, the resultant setting can be termed as \textit{partially transductive}. To solve this, we propose a pipeline that uses an unlearning technique to obtain true inductive baselines. In this new inductive setting, the methods show a significant drop in performance (\textbf{-55\%} on average among 13 baselines with multiple datasets). We validate the unlearning technique using oracle baselines.
An improved few-shot classification technique is proposed that consistently obtains state-of-the-art performance over 13 other recent baseline methods on a comprehensive analysis with 5880 experiments - varying the datasets, differing number of few-shot examples, unlearning setting, and with different seeds. Thus, we identify the issue with the evaluation of CLIP-based few-shot classification, provide a solution using unlearning, propose new benchmarks, and provide an improved method. Our code and models are available \href{https://github.com/akres001/Rethinking-Few-Shot-CLIP-Benchmarks-A-Critical-Analysis-in-the-Inductive-Setting/tree/main}{here}.

\end{abstract}

\section{Introduction}
\label{sec:intro}

Contrastive Language Image Pretraining (CLIP) model \cite{radford_2021_learning} is a widely used vision-language foundational model.
Radford {\it et al.} \cite{radford_2021_learning} demonstrated the ability for CLIP to improve its classification using a few examples (termed linear probing). Subsequently, various works have further improved the ability for a CLIP model to be used as a few-shot classification method. ~\cite{gao_2021_clipadapter, yu_2023_task, kgcoop23, CoPrompt, prograd, zhou_2022_coop, zhou_2022_cocoop, khattak_MaPLe,Khattak_promptsrc, sep}

\begin{figure}[!t]
    \centering
   
     \includegraphics[width=1.\linewidth]{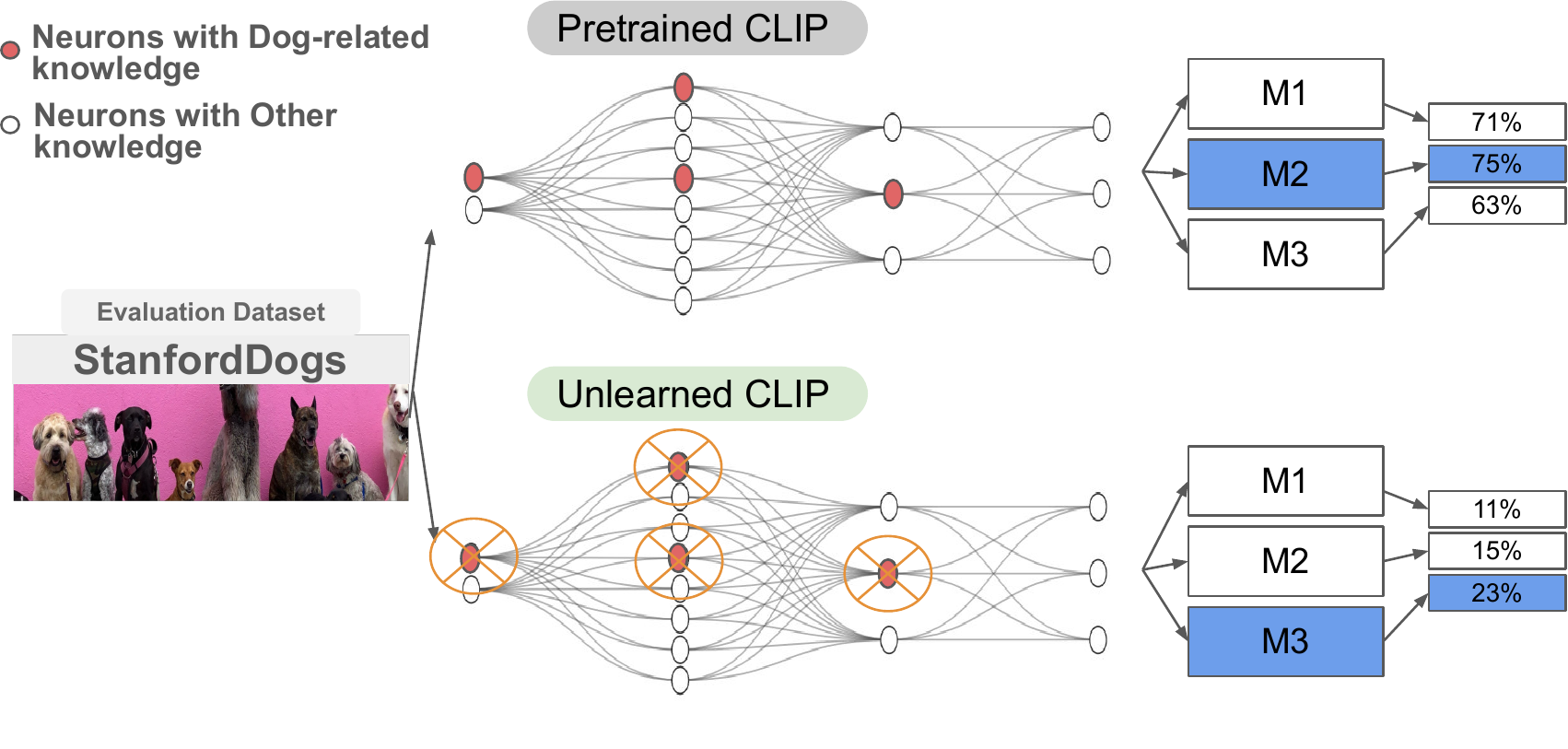}
   
        \caption{ Comparisons between the previous evaluation pipeline (Top: partially transductive) and the proposed pipeline (Bottom: inductive). Top: The tested methods ($M_{1-3}$) apply the pretrained CLIP model which the target classes are likely to be included during training. Bottom: The proposed inductive pipeline aims to provide the test methods with an updated CLIP model with target class information unlearned. The target dataset, unlearning module and few-shot methods are all replaceable.
    }

    \vspace{-3mm}
    \label{fig:teaser}
\end{figure}

The improvement obtained from the few-shot classification is crucial as it can benefit the classification tasks with limited samples. Existing methods can achieve good results with tasks such as classifying the latest dress styles for which the related knowledge has been learned during pretraining on CLIP. However, there are also lots of practical scenarios that need to classify unseen classes with few-shot techniques,  e.g., when it is used to identify new types of AI-generated harmful content or emerging diseases. 
For these use cases, the methods need to be evaluated using \textit{inductive} benchmarks where the model learns specific information about a class from the few-shot examples of the class. 

\begin{figure}[!t]
   \centering
   \includegraphics[width=0.75\linewidth]{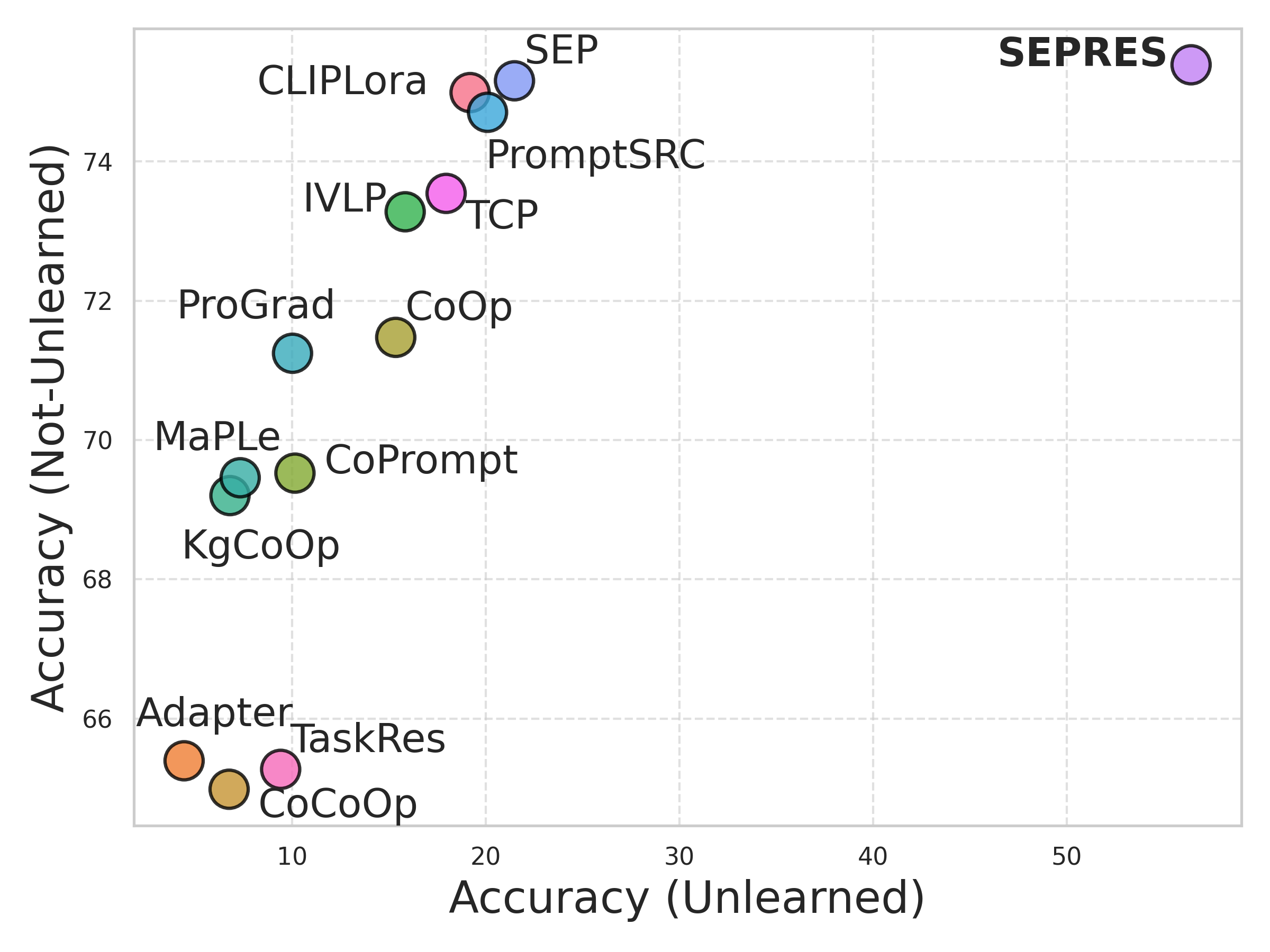}
   
   \caption{Unlearned vs Not Unlearned performance for all few-shot methods. All baseline methods perform much worse in our setting with our proposed SEPRES method being robust in both settings.}
   \label{fig:scatter_notun_un}
   \vspace{-3mm}
\end{figure}

However, the current evaluation of few-shot CLIP classification models is obtained using standard few-shot classification datasets. These results are not indicative of true few-shot generalization performance for unseen classes. As CLIP is trained with a large corpus, most existing datasets would have been seen by the model. Evaluating using a few-shot classification method would result in a \textit{partially transductive} evaluation - where the performance for a specific class is based on previously seen test samples for that specific class. As shown in Fig.~\ref{fig:teaser}, we investigate this aspect of the problem and propose a pipeline for evaluating few-shot methods in an inductive manner. Specifically, we provide a means to fairly evaluate on any benchmark by incorporating unlearning. In this new setting, a significant difference in performance for all 13 CLIP-based few-shot baselines is observed as indicated in Fig. \ref{fig:scatter_notun_un}.

Moreover, one could question whether the unlearning method itself is successful in unlearning a class. E.g., it may be that the unlearning method itself demonstrates poor classification performance, but the knowledge may still be retained within the model. To ensure the validity of the unlearning method, we train two types of CLIP models from scratch, one using the data from all the classes in the ImageNet dataset and another set of CLIP models trained on subsets of ImageNet that exclude certain classes.
In this way, we can provide an \textit{oracle} baseline where the model has not seen the class data as a fair baseline. The trend in few-shot classification performance is similar in the oracle baseline and the corresponding unlearned baselines.

The methodology we propose is flexible as it can be used to create an inductive benchmark out of any dataset and any model for which a reliable unlearning method exists, i.e., an unlearning method that can retain most of the model's original knowledge after removing the information about a specific benchmark dataset.

In this paper, a baseline CLIP-based few-shot learning method is also proposed. Unlike previous methods which tend to modify the weights towards the end of the CLIP model~\cite{yu_2023_task} or modify the weights internally based on the latent space prompt \cite{sep}, the proposed method called Self-Enhanced Prompt Tuning with Residual Textual Features (SEPRES) tends to modify weights both internally and towards the end of the model simultaneously. As shown in Fig.~\ref{fig:scatter_notun_un} and other experimental results in Sec.~\ref{sec:results}, SEPRES consistently achieves the best performance against other baselines in various settings. 

\noindent We make the following \textbf{contributions:} 
\begin{itemize}
    \item A problem with the few-shot CLIP classification is identified. Existing few-shot CLIP classification methods are not evaluated using inductive benchmarks. The usual high performance for most methods is due to the partially transductive nature of evaluation (as shown in Fig. \ref{fig:scatter_notun_un}). 
    \item A novel pipeline is proposed to obtain inductive benchmarks using unlearning - as shown in Tab. \ref{tab:few_shot_results_avgs}. Accuracy of methods like CoOp \cite{zhou_2022_coop} drops sharply from 71.4\% to just 15.3\% on average across datasets and shots in the proposed pipeline for inductive evaluation. A detailed analysis with all components under different setups of the proposed pipeline is conducted with \textbf{5880} experiments as shown in Fig. \ref{fig:aggressive_unlearning}. 
    \item We validate the unlearning method using oracle baselines. The results in Fig. \ref{fig:scratch_methods_nosepres}  show that the proposed solution is consistent with oracle baselines. 
    \item A baseline few-shot learning method, SEPRES, is proposed, which achieves state-of-the-art results against all 13 baselines under different settings and datasets. 
\end{itemize}

\section{Related Work}
\label{sec:relwork}

\paragraph{Few-shot learning in CLIP}

In the field of CLIP-based few-shot learning, prompt tuning has emerged as a key strategy to improve generalization and adaptability to novel tasks. Traditional prompt engineering involves manually crafting prompts, but recent methods \cite{zhou_2022_coop, zhou_2022_cocoop, prograd, kgcoop23} automate this process by learning continuous prompts. Later methods \cite{khattak_MaPLe, Khattak_promptsrc, Khattak_promptsrc, TCP24, sep} expanded to multimodal prompting by jointly optimising text and image prompts and enhancing visual-linguistic synergy. Beyond prompt-based methods, adapters \cite{gao_2021_clipadapter} offer another approach by introducing learnable parameters that adjust pre-trained features without disrupting the model's original knowledge. \cite{CoPrompt} integrates prompting with adapter tuning methods. \cite{cliplora} fine-tunes CLIP's original weights. Training-free adapters \cite{kravets2024clipadaptationintramodaloverlap,zhang_2022_tipadapter,udandarao_2023_susx} have also been proposed but they rely on CLIP's inherent zero-shot accuracy for their effectiveness and are unable to add new knowledge. Since our focus is on evaluating few-shot adaptation where CLIP’s zero-shot accuracy is limited, we exclude such methods from our evaluation scope. Our focus in this paper is to thoroughly investigate the generalization performance of these methods, especially in the inductive setting that we obtain using unlearning. We also provide a baseline method that improves performance over these methods in the inductive setting.

% \vspace{-0.5cm}
\paragraph{Machine unlearning}
Unlearning in multimodal models such as CLIP is an emerging area, particularly due to the complexity of handling interactions between visual and textual modalities. To the best of our knowledge there exist only two methods that address unlearning in CLIP. Authors in \cite{clipkrav} utilize Lipszhitz regularization to smooth both visual and textual embeddings associated with the class intended to be forgotten relative to the perturbation introduced to the samples from that class. Authors in \cite{clipkravtextproj} remove the desired classes by adjusting the text projection matrix via low-rank adaptation. Both of these methods are zero-shot and do not require any visual examples. Among non-zero-shot methods, SalUn \cite{fan2023salun} uses weight saliency to unlearn the desired data. Unrolling SGD \cite{sgdunl} reverses gradient updates to forget data points but relies on models trained with SGD, unlike CLIP, which uses Adam. Another approach, \cite{WarPirWreRie20}, leverages influence functions to remove specific features and labels. SSD \cite{Foster_Schoepf_Brintrup_2024} updates only the parameters relevant to the forget but not to the retain data to achieve unlearning. To our knowledge, these non-zero-shot unlearning methods have not been applied to CLIP. In this paper, we need to unlearn a lot of classes and we find that zero-shot unlearning methods over-forget as the number of classes becomes very high. Hence, we adapt an existing non-zero shot machine unlearning method to forget desired classes. We compare  various unlearning methods to justify our choice in the Supplementary material. Note that a better unlearning technique can always be adopted in our proposed framework.

% \section{Towards Inductive Few-shot CLIP}
\section{Partially Transductive $\rightarrow$ Inductive}
\label{sec:clipbackground}

\subsection{Inductive Few-shot Learning}

In standard few-shot learning setting ($m$-way $k$-shot)~\cite{hu2022pmf, vinyals2016matching, chen2021self}, we have \( \mathcal{C}_{\text{train}} \): set of base classes used for initial training, \( \mathcal{C}_{\text{novel}} \): set of novel $m$ classes ($m$-way) which is unseen during training, \( D_{train} = \{(x_i, y_i)\}_{i=1}^{N} \): training set with sample $x_i$ from base classes \( y_i \in \mathcal{C}_{\text{base}} \), \( D_{Fs} = \{(x_i, y_i)\}_{i=1}^{k*m} \): support set with \( k \) labeled samples ($k$-shot) per novel class \( y_i \in \mathcal{C}_{\text{novel}} \) and \( D_{Fq} = \{(x_j, y_j)\}_{j=1}^{q} \): query set with \( q \) labeled samples with \( y_j \in \mathcal{C}_{\text{novel}} \) for testing.  $f$ is the model trained on $D_{train}$.

Through fine-tuning~\cite{chen2019closer}, a loss (e.g. cross-entropy) is then minimized using few-shot examples with adapted parameters on the original model $f$ that we denote $\hat{f}$:
\begin{equation}
    \mathcal{L}_{\text{few-shot}} = \frac{1}{m} \sum_{(x_j, y_j) \in D_{Fs}} \ell(\hat{f}(x_j; f), y_j).
\label{eq:standard_fewshot_loss}
\end{equation}
This final model is tested on the query set $D_{Fq}$. 

\subsection{Few-shot Classification with CLIP}

CLIP learns to align image features with corresponding text representations in a shared embedding space with contrastive pre-training on the dataset $D_{train}^{C} = \{(x_i, t_i)\}_{i=1}^{\mathcal{N}}$ 
% \begin{equation}
%     D_{train}^{C} = \{(x_i, t_i)\}_{i=1}^{\mathcal{N}}
% \label{eq:clip_data_ori}
% \end{equation}
where $\{x_i, t_i\}$ is the image sample and its corresponding text from label $y_i$, $\mathcal{N}$ is the number of pairs. Similar to most of the pre-trained foundation models, the details of this dataset are unknown to us.

For zero-shot classification with CLIP across $m$ classes ($\mathcal{C}_{\text{novel}}$), each class name is inserted into a prompt like ``A photo of a \{class\}" and embedded using CLIP text encoder. This forms the classifier's weight matrix $W \in \mathbb{R}^{m\times d}$, where \textit{d} is the embedding dimension. Similarly, given a test image, it is encoded using the CLIP image encoder obtaining $F \in\mathbb{R}^{d}$ image embedding. Next, the dot product of $W$ and $F$ produces classification logits for the image:
\begin{equation}
    \text{Logits} = F W^T, \ \text{Logits} \in  \mathbb{R}^{m}.
\label{eq:clip_logits}
\end{equation}
When the support set $D_{Fs}$ is provided with $k$ labeled samples for each class in $\mathcal{C}_{\text{novel}}$, the CLIP-based zero-shot task is naturally transferred to a $m$-way $k$-shot few-shot learning task with CLIP. A similar procedure can be done as shown in Eq.~\ref{eq:standard_fewshot_loss}, keeping the original CLIP frozen while optimizing additional adapter parameters. In this paper, 13 CLIP-based few-shot methods are compared as detailed in Sec.~\ref{sec:unbenres}.

\subsection{Partially Transductive vs. Inductive}

When comparing few-shot learning w/wo large foundation models like CLIP, a key difference between them is the control over the base training set $D_{train}$ and $D_{train}^{C}$. The former is easy to control and make sure $D_{train} \cap D_{Fs} = \varnothing$, so that the unseen classes (i.e., $\mathcal{C}_{\text{base}} \cap \mathcal{C}_{\text{novel}}  = \varnothing$) can be tested and all few-shot learning methods can be inductively evaluated. Large models like CLIP, however, leverage extensive, diverse datasets from different resources, which are concluded as $D_{train}^{C}$ in our paper. As the evaluation datasets for few-shot learning such as StanfordCars ~\cite{krause_2013_3d}, CUB ~\cite{he_2020_finegrained}, etc. are commonly used in various computer vision tasks, it is almost certain that $D_{train}^{C} \cap D_{Fs} \neq \varnothing$. Hence, there is an index $i$ such that for most labels $y_j$ in the support set $D_{Fs}$, a highly correlated text $t_i$ can be found and paired. In this paper, we define it as \textit{partially transductive}.

True generalization for a few-shot method on unseen classes that have not been seen can only be understood by evaluating in the inductive setting. We aim to solve this problem using the proposed pipeline which enables us to use large models for few-shot learning tasks in an inductive manner. Our framework also enables constructing explicit partially transductive benchmarks in an explicit manner. Note that theoretical bounds as obtained in \cite{Galanti2022GeneralizationBF} are not possible due to the unavailability of the training data.

\section{Method}
\label{sec:method}

The proposed pipeline for generating inductive benchmark datasets is shown in Fig.~\ref{fig:teaser} (Bottom). This pipeline relies on a model, an unlearning method and a standard evaluation dataset. First, unlearning is performed on the given CLIP model removing the information about the classes in the dataset making this dataset a new inductive benchmark for CLIP. The unlearned CLIP model is then applied to different few-shot learning methods on the new benchmark. This procedure allows for a comprehensive evaluation of performance in an inductive manner. The proposed pipeline can be generally applied for different tasks/methods as most of the components in this pipeline are replaceable.

\subsection{Unlearning CLIP}

Retraining CLIP from scratch on its original dataset, excluding specific classes to evaluate the models in an inductive manner, is not feasible due to computational constraints and the lack of public access to the original dataset $D_{train}^{C}$. Instead, unlearning is applied as a practical alternative. While zero-shot unlearning methods~\cite{clipkrav, clipkravtextproj} were proposed for CLIP, it is noticed that these approaches tend to overly diminish the knowledge we aim to retain when unlearning a large number of classes. Other unlearning techniques, such as~\cite{fan2023salun, graves_2020_amnesiac}, either inadvertently erase non-targeted knowledge or struggle to effectively unlearn the target information. Among the evaluated methods, Selective Synaptic Dampening (SSD)~\cite{Foster_Schoepf_Brintrup_2024}, which we adapted for CLIP, is the most effective for preserving non-targeted knowledge while achieving robust unlearning. SSD performs unlearning by selectively updating parameters that are relevant to the ``forget set" but not to the ``retain set". This is estimated via the Fisher Information Matrix (FIM). To compute the FIM for CLIP, the logits for each image are computed, as shown in Eq.~\ref{eq:clip_logits}. Then, given the one-hot encoded vector with the true label, we compute the cross-entropy loss given by:

\begin{equation}
    \mathcal{L} = -\sum_{i=1}^{Cls} y_i log(s_i),
\end{equation}
where $Cls$ is the total number of classes, $s_i$ is the logit for class $i$ and $y_i$ is 1 for correct class, 0 otherwise. FIM is then computed using this loss as follows:
\begin{equation}
\mathcal{F}(\theta) = \mathbb{E}_{(x, y) \sim p(x, y)} \left[ \nabla_{\theta} \mathcal{L}(x,y;\theta) \nabla_{\theta} \mathcal{L}(x,y;\theta)^\top \right].
\end{equation}

We compute FIMs over the forget $D_f$ and the retain $D_r$ sets, denoted as $\mathcal{F}_{f}$ and $\mathcal{F}_{r}$ respectively. Then, each parameter of the model $f_{\theta}$ is updated as follows:
\begin{equation}
\theta_i = 
\begin{cases} 
\beta \theta_i & \text{if } \mathcal{F}_{f,i} > \alpha \mathcal{F}_{r,i} \\
\theta_i & \text{if } \mathcal{F}_{f,i} \leq \alpha \mathcal{F}_{r,i}
\end{cases}
\quad \forall i \in \{1, ..., |\theta|\},
\end{equation}
\begin{equation}
\beta = \min\left( \frac{\lambda \mathcal{F}_{r,i}}{\mathcal{F}_{f,i}}, 1 \right), 
\end{equation}

where $\alpha$ is the hyperparameter controlling how selective the pruning should be, $\beta$ is the dampening factor, computed based on the relative importance of the parameter in the forget and the retain sets, bounded by $1$, and $\lambda$ controls the strength of the dampening. Note that $\beta \to 0$ for highly specialized parameters for the forget set. This allows parameters that are important for the forget set to be reduced in magnitude without fully pruning them, protecting general parameters that are important for both forget and retain sets.

Even though the method requires real retain data we find that we do not need to provide all the data that CLIP has seen during pretraining but only a few datasets, which is enough to keep original CLIP's knowledge almost untouched as validated on some held-out validation datasets as shown in Tab.~\ref{tab:forget_results}. We show ablations with other unlearning methods in the Supplementary material (Section IV).

With the proposed pipeline, every dataset can be made a benchmark for an inductive test. As presented in the results section - row 1 of Tab.~\ref{tab:few_shot_results_avgs}, the zero-shot accuracy of CLIP is now very low on the testing datasets while other knowledge is preserved (Tab.\ref{tab:forget_results}). Each dataset can be used as a benchmark to test few-shot methods with CLIP inductively as if CLIP has never seen the classes from the benchmark.

\begin{figure*}
        \centering
        \includegraphics[width=0.84\linewidth]{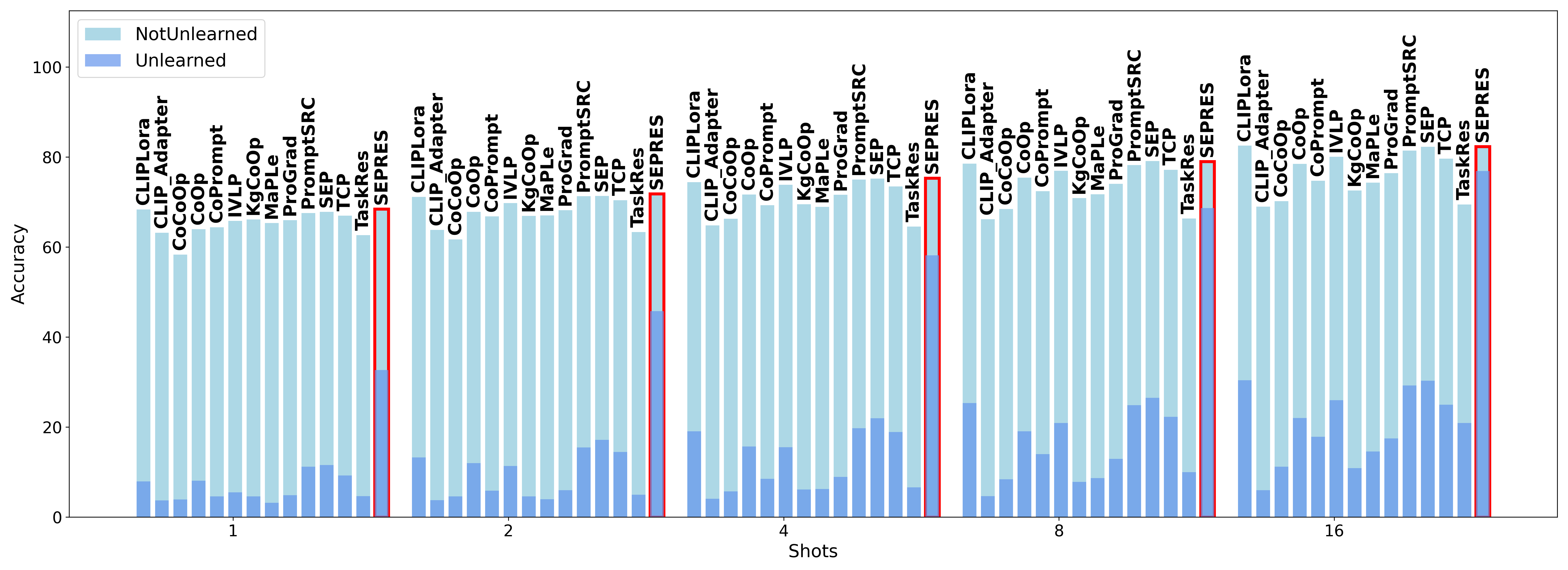}
        \caption{Accuracy of different few-shot learning methods aggregated across 7 datasets and different shots for CLIP without unlearning vs. unlearned CLIP. Compared to the transductive setting (NotUnlearned), the performance of the few-shot baselines has shown a clear gap when compared with the performance in the inductive setting (Unlearned). While SEPRES performs consistently better (in red blocks), even if the gap remains big in low-shot regimes such as 1,2 and 4 shots. }
        \label{fig:acc_un_notun}

\vspace{-2mm}
\end{figure*}

\begin{table*}[t!]
\centering
\resizebox{0.98\textwidth}{!}{\begin{tabular}{|c|c||c|c|c|c|c|c|c|c|c|c|c|c|c|c|}
\hline
\rowcolor{gray!20} \textbf{Dataset} & \textbf{Shots} & \textbf{CLIPLora} & \textbf{CLIPAdapter} & \textbf{CoCoOp} & \textbf{CoOp} & \textbf{CoPrompt} & \textbf{IVLP} & \textbf{KgCoOp} & \textbf{MaPLe} & \textbf{ProGrad} & \textbf{PromptSRC} & \textbf{SEP} & \textbf{TCP} & \textbf{TaskRes} & \textbf{SEPRES} \\ \hline
Average & 0 & 0.035 & 0.035 & 0.035 & 0.035 & 0.035 & 0.035 & 0.035 & 0.035 & 0.035 & 0.035 & 0.035 & 0.035 & 0.035 & 0.035 \\ \hline
Average & 1 & 7.936 & 3.719 & 3.895 & 8.074 & 4.605 & 5.500 & 4.582 & 3.198 & 4.840 & 11.195 & 11.571 & 9.240 & 4.657 & \textbf{32.662} \\ \hline
Average & 2 & 13.262 & 3.781 & 4.590 & 11.976 & 5.855 & 11.336 & 4.597 & 3.952 & 5.950 & 15.481 & 17.133 & 14.448 & 4.957 & \textbf{45.757} \\ \hline
Average & 4 & 19.040 & 4.048 & 5.705 & 15.707 & 8.493 & 15.514 & 6.123 & 6.257 & 8.933 & 19.757 & 21.948 & 18.924 & 6.586 & \textbf{58.162} \\ \hline
Average & 8 & 25.349 & 4.629 & 8.393 & 19.062 & 13.976 & 20.924 & 7.812 & 8.671 & 12.941 & 24.848 & 26.500 & 22.252 & 9.986 & \textbf{68.686} \\ \hline
Average & 16 & 30.414 & 5.990 & 11.207 & 21.990 & 17.857 & 25.971 & 10.896 & 14.567 & 17.496 & 29.224 & 30.310 & 24.950 & 20.910 & \textbf{76.871} \\ \hline
Overall Average (excl. zs) & - & 19.200 & 4.433 & 6.758 & 15.362 & 10.157 & 15.849 & 6.802 & 7.329 & 10.032 & 20.101 & 21.492 & 17.963 & 9.419 & \textbf{56.428} \\ \hline
\end{tabular}}
\caption{Inductive setting results averaged across 7 datasets and different shots. SEPRES performs best in this setting, while other few-shot methods are unable to generalize from few-shot examples when the knowledge about the benchmark dataset classes is removed from CLIP.}
\label{tab:few_shot_results_avgs}
\vspace{-2mm}
\end{table*}

\subsection{Knowledge Lost During Unlearning}
\label{sec:knowlost}
As a proxy for the total knowledge lost of the CLIP model after unlearning a targeted dataset, we measure the accuracy reduction on some validation datasets (Sec.~\ref{sec:datasets}) that are never used as retain sets. A good unlearning method should have low accuracy on the target set while preserving not targeted knowledge. While we select validation sets with non-overlapping classes, semantic similarities may exist. For instance, unlearning UCF101 \cite{soomro_2012_ucf101}, we may observe a greater loss of knowledge when evaluating the remaining accuracy on Caltech101 \cite{feifei_2007_learning} because certain classes, such as ``biking" in UCF101 and ``motorbike" in Caltech101, are semantically similar and are therefore affected by the unlearning process.

To account for this, we compute a weighted average of the accuracy reduction across validation sets, where weights reflect the similarity between the validation and the unlearned datasets. Validation sets more similar to the unlearned dataset contribute less to the total knowledge loss estimate. We can measure this similarity using four metrics: Maximum Mean Discrepancy (MMD) \cite{mmd} computed separately on image and text embeddings, then averaged to produce a unified similarity score; WordNet distance between textual classes that approximates the task-space geometry approach from \cite{taskspace} to compute tasks similarities; Proxy-A distance (PAD) \cite{bendavid_2006_analysis} computed on image embeddings; and uniform weighting that provides equal weighting and does not take into account the class similarities. 
% As shown in Tab.~\ref{tab:forget_results}, all metrics produce similar results with low knowledge lost on the validation sets after unlearning. Based on these comparable outcomes, we adopt MMD weighting as our primary metric throughout this paper.
The total knowledge loss (TKL) of CLIP is computed as follows:
\begin{align}
    \text{TKL} = \sum_{i \in V} w_i \times \text{KnowledgeLost}_i \label{eq:lostacc_mmd} 
    % \sum_{i \in V} w_i = 1, \hspace{15.5em} 
\end{align}
where $KnowledgeLost_i$ is the accuracy degradation on the validation dataset $i$ after performing unlearning, $V$ is the collection of validation datasets and $w_i$ (with $\sum_{i \in V} w_i$ = 1) is the normalized weight computed using MMD, PAD, WordNet distance or uniform weighting schemes. 

% Maximum Mean Discrepancy (MMD) \cite{mmd} is obtained with RBF kernel metrics between validation and unlearned dataset for both images and textual class embeddings, and can be used as weights. MMD with RBF kernel is bound between 0 and 1 and is 0 when the two distributions are identical while closer to 
% 1 when they are maximally different.
% $\text{MMD}_{\text{t}}(D_{\text{u}}, D_i)$ is the MMD computed between the textual class embeddings of the unlearned dataset $D_u$ and and the validation dataset $D_i$. Similarly for the images $\text{MMD}_{\text{im}}$.
% w_i &= \frac{1}{2} \left( \text{MMD}_{\text{t}}(D_{\text{u}}, D_i) + \text{MMD}_{\text{im}}(D_{\text{u}}, D_i) \right) \\

\subsection{Self-Enhanced Prompt Tuning with Residual Textual Features (SEPRES)}
In addition to the existing CLIP-based few-shot learning methods, a new baseline, SEPRES (Self-Enhanced Prompt Tuning with Residual Textual Features), is proposed. This is based on the previous Self-Enhanced Prompt Tuning (SEP) \citep{sep} method. 
SEP integrates knowledge from frozen tokens into new learnable prompts using a Token-Fusion Module (TFM) to combine pre-trained visual or textual tokens with these learnable prompt-related tokens. This fusion, as the authors claim, enhances both the discriminative and generative capabilities of the tokens, improving CLIP's generalization across seen and unseen domains. However, while SEP is highly effective at extracting knowledge already embedded in CLIP, the fusion process integrates too much pre-trained information into the learnable prompts. This restricts the method's capacity to learn new knowledge not seen during pretraining, which is essential for generalizing to unseen domains. Adding additional learnable prompts inside the model that are passed through the TFM module would still limit the learning of new knowledge. This is because these parameters would pass through CLIP’s internal layers and would be forced to conform to CLIP’s existing representations rather than learning new representations. 
To overcome these limitations and improve generalization, we propose adding learnable parameters in a residual fashion to the final textual embeddings, inspired by the approach in TaskRes. This design allows the model to retain the benefits of CLIP’s pretrained knowledge while learning new knowledge. For example, when adapting to dogs in few-shot learning, the method can leverage its understanding of other animals (via SEP) while learning new knowledge (via RES) not seen during pretraining from few-shot examples of dogs. Based on the results in Sec.~\ref{sec:unbenres}, the proposed method SEPRES achieves state-of-the-art results across all tested datasets in an inductive manner. Note that the main focus of this work is creating the inductive benchmark, and the proposed baseline method suggests that using this insight, one can create better generalised few-shot methods that work inductively as well as transductively as shown. The mathematical details of SEPRES are included in the Supplementary Material (Section II).

\begin{figure*}
    \centering
    \begin{subfigure}[t]{0.67\textwidth}
        \centering
        \includegraphics[width=1.\linewidth]{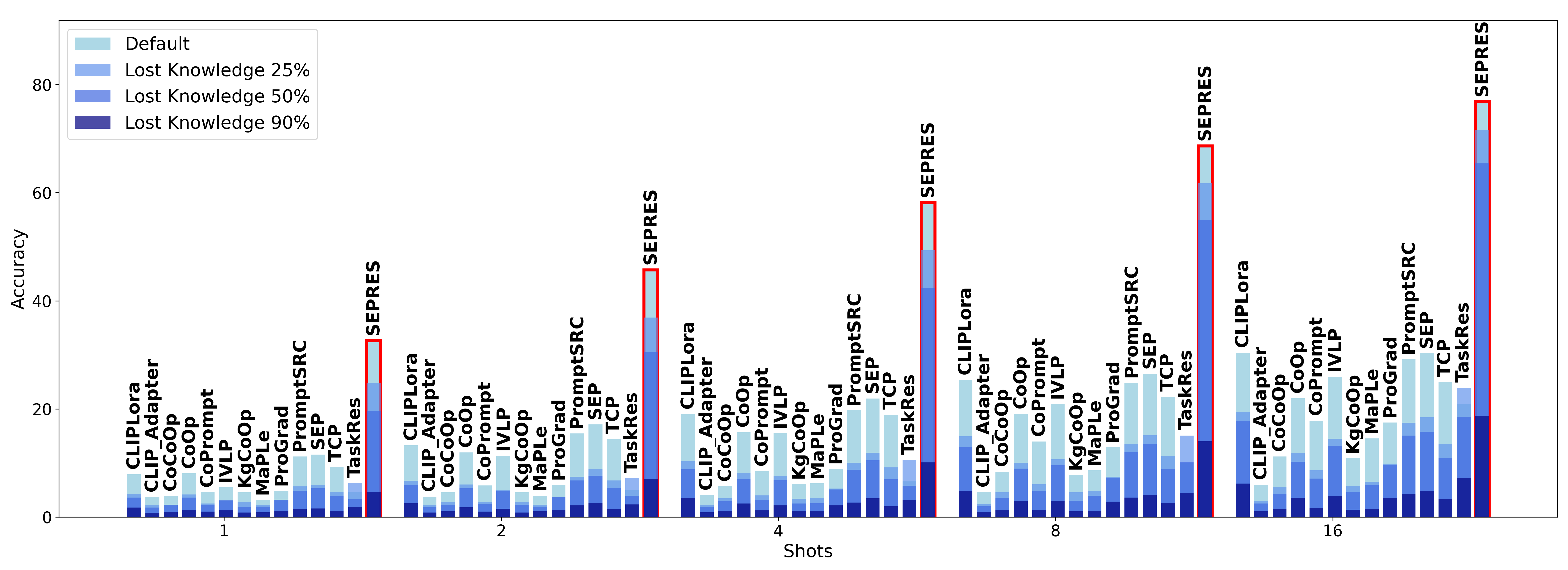}
        \caption{}
        \label{fig:aggressive_unlearning}
    \end{subfigure}%
    \hfill
    \begin{subfigure}[t]{0.3\textwidth}
        \centering
        \vspace{-115pt} % Adjusts vertical space
        % \begin{table}[!b]
        % \centering
        \resizebox{1.\textwidth}{!}{
        \begin{tabular}{|c|p{2.2cm}|c|c|c|c|}
        \hline
        % \rowcolor{gray!20} \textbf{Forget Ds} & \textbf{Retain Ds} & \textbf{WordNet} & \textbf{MMD} & \textbf{Uniform} & \textbf{PAD} \\ \hline
        \rowcolor{gray!20} \textbf{Forget Ds} & \textbf{Retain Ds} & \multicolumn{4}{c|}{\textbf{Total Knowledge Lost with weights from:}} \\ \cline{3-6}
        \rowcolor{gray!20} & & \textbf{WordNet} & \textbf{MMD} & \textbf{Uniform} & \textbf{PAD} \\ \hline
        SDogs & C101, OxFlowers, CUB & 2.293 & 2.191 & 2.265 & 2.240 \\ \hline
        SCars & C101, CUB, FGVCA & 2.196 & 2.312 & 2.203 & 2.335 \\ \hline
        C101 & SDogs, OxFlowers, UCF101, FGVCA & 2.174 & 1.793 & 2.148 & 1.872 \\ \hline
        FGVCA & SCars, C101, OxFlowers & 0.593 & 0.636 & 0.640 & 0.641 \\ \hline
        UCF101 & SDogs, C101, CUB & 2.265 & 1.230 & 2.253 & 1.913 \\ \hline
        OxFlowers & SDogs, C101, CUB & 2.893 & 2.756 & 3.206 & 2.644 \\ \hline
        CUB & C101, OxFlowers, UCF101 & 1.880 & 1.727 & 1.832 & 1.904 \\ \hline
        \end{tabular}
        }
        \caption{}
        \label{tab:forget_results}
    \end{subfigure}
    \caption{(a) Shows few-shot accuracy of different methods aggregated across 7 datasets and all shots for various levels of CLIP's lost knowledge. \textit{Default} represents the minimum level of lost knowledge to forget the dataset as shown in (b). As more knowledge is lost, the accuracy across all methods goes down, with SEPRES demonstrating the best robustness. (b) Shows the retain datasets used while unlearning a dataset and total knowledge lost (Eq.\ref{eq:lostacc_mmd}) on the validation sets using different metrics to weigh the similarity between the validation and unlearned set. Across all datasets, the general knowledge is almost fully preserved after unlearning. }

\end{figure*}
\vspace{-2mm}
\section{Results}
\label{sec:results}

\subsection{Datasets}
\label{sec:datasets}

In this paper, experiments are conducted on the following datasets that have non-overlapping classes to avoid the interference of overlapping classes during unlearning and testing:
Caltech101 ~\cite{feifei_2007_learning}, EuroSAT 
~\cite{helber_2019_eurosat}, StanfordCars ~\cite{krause_2013_3d}, DescribableTextures ~\cite{cimpoi_2013_describing}, OxfordFlowers 
~\cite{nilsback_2008_automated}, Food101 ~\cite{bossard_2014_food101}, FGVCAircraft ~\cite{maji_2013_finegrained}, StanfordDogs ~\cite{khosla_novel}, PLANTDOC ~\cite{singh_2020_plantdoc}, CUB ~\cite{he_2020_finegrained}, UCF101 \cite{soomro_2012_ucf101} and SUN397 \cite{xiao_2010_sun}.

For unlearning of these 12 datasets, we use 5 of them (i.e., Food101, EuroSAT, DescribableTextures, SUN397 and PLANTDOC) for validation that are only used to assess the remaining knowledge of CLIP after unlearning. These datasets are not used for unlearning nor for retaining the information. Full training set is used to estimate parameters sensitive to the knowledge retention and to unlearn a specific dataset.
For the remaining 7 datasets we apply unlearning one dataset at a time. For instance, the StanfordDogs is unlearned from the CLIP model. Then CLIP-based few-shot methods can be evaluated with the unlearned CLIP model using few-shot examples from StanfordDogs. To validate that the CLIP model's performance has not changed on other datasets, the accuracy of the five held-out validation datasets becomes the proxy for CLIP's general knowledge retention. The same experiments for each of the 7 datasets have been done on all the 14 methods, including the proposed SEPRES method. When unlearning a dataset, the method requires real retain data to ensure that the knowledge of CLIP is retained. This can be achieved using the samples from the remaining 6 datasets. For each forget/target dataset, the datasets used for knowledge retention are provided in Tab. \ref{tab:forget_results}. This retain set ensures that the CLIP's total knowledge is fully retained as validated using the metric in Eq. \ref{eq:lostacc_mmd} on the validation datasets. 

\begin{figure*}[!t]
    \centering
    \includegraphics[width=1.\linewidth]{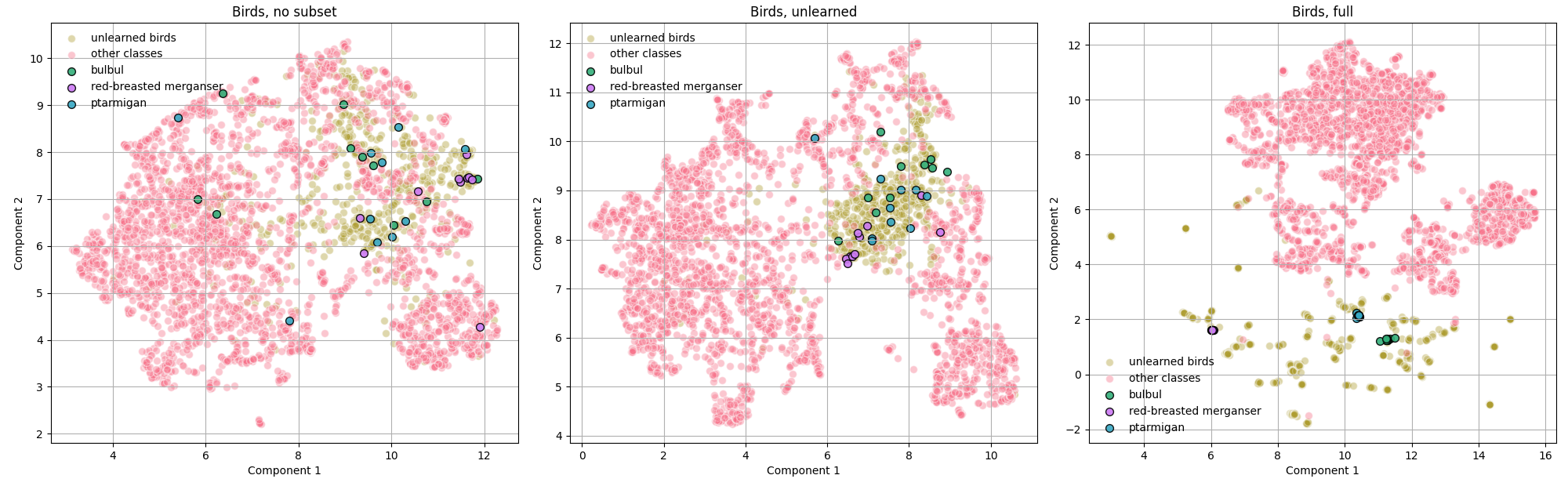}
    
    \caption{Visual features distributions across different settings with CLIP trained from scratch. Left visualisation: excluding the subset from training. Middle visualisation: unlearning the subset. Right visualisation: without any unlearning. Classes to unlearn are shown in light green, while other classes are in light red. \textit{No subset} and \textit{unlearned} settings with the highlighted classes from the excluded subset are more sparse and overlapping compared to the \textit{full} setting. The unlearned subset in both \textit{no subset} and \textit{unlearned} settings overlap much more with other classes compared to the full setting. All indicate that \textit{no subset} and \textit{unlearned} settings are similar. }

    \label{fig:umap_scratch_wlabels}
    \vspace{-2mm}
\end{figure*}

\begin{table*}[!t]
\centering
\resizebox{1.\textwidth}{!}{\begin{tabular}{|c|c||c|c|c|c|c|c|c|c|c|c|c|c|c|c|}
\hline
\rowcolor{gray!20} \textbf{Dataset} & \textbf{Type} & \textbf{CLIPLora} & \textbf{CLIPAdapter} & \textbf{CoCoOp} & \textbf{CoOp} & \textbf{CoPrompt} & \textbf{IVLP} & \textbf{KgCoOp} & \textbf{MaPLe} & \textbf{ProGrad} & \textbf{PromptSRC} & \textbf{SEP} & \textbf{TCP} & \textbf{TaskRes} & \textbf{SEPRES} \\ \hline
Dogs & \textcolor{orange}{Excluded} & 4.942 & 5.727 & 1.700 & 2.593 & 6.047 & 5.453 & 6.479 & 5.553 & 2.831 & 5.587 & 8.287 & 6.487 & 6.713 & \textbf{9.660} \\ \hline
Birds & \textcolor{orange}{Excluded} & 9.754 & 7.867 & 2.193 & 1.900 & 7.313 & 7.730 & 8.127 & 6.207 & 2.662 & 9.163 & 13.313 & 9.613 & 8.333 & \textbf{17.507} \\ \hline
Vehicles & \textcolor{orange}{Excluded}  & 9.900 & 7.667 & 2.253 & 4.273 & 6.047 & 7.395 & 8.415 & 6.440 & 3.413 & 9.173 & 12.273 & 10.993 & 9.067 & \textbf{19.853} \\ \hline
\hline
Dogs & \textcolor{teal}{Unlearned} & 6.914 & 4.820 & 4.193 & 4.440 & 5.080 & 5.533 & 4.431 & 4.673 & 2.825 & 6.293 & 6.487 & 7.793 & 8.653 & \textbf{16.953} \\ \hline
Birds & \textcolor{teal}{Unlearned} & 13.756 & 8.220 & 6.127 & 4.673 & 8.147 & 9.447 & 5.127 & 7.127 & 3.029 & 11.753 & 12.327 & 13.567 & 9.020 & \textbf{34.713} \\ \hline
Vehicles & \textcolor{teal}{Unlearned} & 8.927 & 6.333 & 4.740 & 5.607 & 6.500 & 7.040 & 6.003 & 6.093 & 2.571 & 8.313 & 7.873 & 9.047 & 9.147 & \textbf{24.200} \\ \hline
\end{tabular}}

\caption{Few-shot methods results with CLIP trained from scratch on ImageNet \textcolor{orange}{excluding the subset} and \textcolor{teal}{unlearning} the subset. Averaged across all shots. The results of excluding a subset from training and unlearning that subset after training achieve similar results across different few-shot learning methods, demonstrating that the two settings are similar from this perspective. }
\label{tab:scratch_subsets_avg}
\vspace{-2mm}
\end{table*}

\subsection{Unlearning Benchmark Results}
\label{sec:unbenres}
In this paper, 13 CLIP-based few-shot learning methods are included: CLIP Adapter \cite{gao_2021_clipadapter}, CoOp \cite{zhou_2022_coop}, CoCoOp \cite{zhou_2022_cocoop}, KGCoOp \cite{kgcoop23}, ProGrad \cite{prograd}, TaskRes \cite{yu_2023_task}, MaPLe \cite{khattak_MaPLe}, PromptSRC \cite{Khattak_promptsrc}, IVLP \cite{Khattak_promptsrc}, TCP \cite{TCP24}, CoPrompt \cite{CoPrompt}, SEP \cite{sep} and LoRA for CLIP \cite{cliplora}. Fig.\ref{fig:acc_un_notun} shows the results of all the methods including SEPRES. They are aggregated across datasets and shots for original and unlearned CLIP. The performance of few-shot methods significantly declines after unlearning in our inductive setting. For example, CoOp is only able to achieve 15.3\% accuracy across different datasets and shots in the unlearned/inductive setting while it was 71.4\% (\textbf{-56.1\%}) in a partially transductive setting. PromptSRC reduced from 74.7\% to 20.1\% (\textbf{-53.6\%}). CLIPLora reduced from 74.9\% to 19.2\% (\textbf{-55.7\%}). On the contrary, SEPRES reduced from 75.4\% to 56\% (\textbf{-19.4\%}) - comparatively a relatively modest difference. More details are shown in the Supplementary (Sections VIII and XII). 

Under the partially transductive setting, the original performance of the baseline methods was based on specific knowledge related to the target classes that CLIP already contained. However, when testing these methods in an inductive setting most of the methods struggle to get high accuracy. As shown in Fig.~\ref{fig:acc_un_notun},  clear performance differences can be observed between transductive and inductive settings on the test sets. 
In contrast, the proposed SEPRES method can generalize well even when relevant knowledge is unlearned, showing a smaller performance gap. Note that the figure provides the average accuracies with different shots on the test sets of the target datasets. When comparing the average accuracy across shots and datasets for unlearned and not unlearned CLIP, as shown in the scatter plot in Fig. \ref{fig:scatter_notun_un}, SEPRES consistently performs well in both conditions as it manages to maintain a good accuracy regardless of whether CLIP has seen classes from the dataset or not. The difference in values between the X-axis for the unlearned CLIP that tends to be lower as compared to the values in the Y-axis for the original CLIP clearly illustrates the gap in performance for the methods. This result motivates the need to consider this aspect as is done in the proposed work. 

Also, we notice that there are some flips in the rankings of the methods in both settings (Fig. \ref{fig:scatter_notun_un}). For example, PromptSRC performs better than CLIPLora in the unlearned setting but underperforms in the normal one. Similarly, TaskRes is better in the unlearned setting than CLIPAdapter, MaPLe and KgCoOp but underperforms them in the normal setting. Thus, testing methods in both settings is important to understand their generalization capabilities. SEPRES is robust in both settings maintaining its ranking.

\begin{table}[!h]
\centering
\small
\resizebox{0.48\textwidth}{!}{
\begin{tabular}{|c|c|c|c|c|c|}
\hline
\multirow{-1}{*}{\cellcolor[gray]{.9}\begin{tabular}[c]{@{}c@{}}\textbf{Dataset}\end{tabular}} & \multicolumn{2}{c|}{\cellcolor{gray!20}\textbf{Unlearning Subset}} & 
\multicolumn{2}{c|}{\cellcolor{gray!20}\textbf{Training w/out Subset}} & 
\multirow{-1}{*}{\cellcolor[gray]{.9}\textbf{MMD ImageNet}} \\ 
\cline{2-5} 
\multirow{-2}{*}{\cellcolor[gray]{.9}} & 
\cellcolor{gray!10}\textbf{Acc. Other Classes} & \cellcolor{gray!10}\textbf{Acc. Subset} & \cellcolor{gray!10}\textbf{Acc. Other Classes} & \cellcolor{gray!10}\textbf{Acc. Subset} & 
\multirow{-1}{*}{\cellcolor[gray]{.9}\textbf{w/o Subset}} \\
\hline
Dogs     & 0.55 & 0.043 & 0.48   & 0.06 & 0.1719 \\ \hline
Birds    & 0.56 & 0.043 & 0.52   & 0.082 & 0.1037 \\ \hline
Vehicles & 0.56 & 0.054 & 0.5346 & 0.07 & 0.1100 \\ \hline
\end{tabular}}
\caption{Accuracy results for unlearning subsets and training without subsets across different datasets, alongside MMD scores between each dataset's subset and ImageNet without that subset.}
\label{tab:scrath_results}
\vspace{-2mm}
\end{table}

Tab. \ref{tab:few_shot_results_avgs} provides an aggregated quantitative summary of the unlearned CLIP results across various shots and methods, averaged over seven datasets. A similar difference in quantitative performance is observed for the 13 baseline few-shot methods when compared against the proposed SEPRES method under the inductive setting. % highlighting the generally poor performance of standard few-shot methods in this setting. 
For more detailed dataset-specific results, please refer to the Supplementary (Section VIII). Fig.~\ref{tab:forget_results} shows the retained datasets that were used during unlearning and total knowledge loss measured with Eq.~\ref{eq:lostacc_mmd}. All weighting metrics (WordNet, MMD, PAD, Uniform - see Sec.~\ref{sec:knowlost} for more details) produce similar total knowledge loss (for most datasets $\leq 2.5\%$) on the validation sets after unlearning. Based on these comparable outcomes, we adopt MMD weighting as our primary metric throughout this paper.

\subsection{Unlearning and Oracle Baselines}

We ask whether the results obtained in the previous section are truly testing the CLIP model in an inductive setting. The only way to test it is to compare these results against a CLIP model trained from scratch excluding the target classes. Training the original CLIP model from scratch is infeasible both for computational reasons and the unavailability of its training data. Thus, to test this, in this paper, CLIP is trained from scratch on ImageNet using contrastive loss. This forms the oracle setting to validate the proposed pipeline. We select three subsets from ImageNet classes related to dogs, vehicles, and birds and train CLIP on all remaining images, excluding each selected subset. 

Our goal is to show that \textit{unlearning} the subset from the CLIP model trained on the full ImageNet has a similar effect to the CLIP model trained from scratch on data excluding the specific selected subset. 

Our training in the oracle setting aims to mimic the existing state-of-the-art unlearning performance \cite{Foster_Schoepf_Brintrup_2024}.
To introduce diversity in textual prompts, we randomly select a template from the ImageNet templates during training \footnote{https://github.com/NVlabs/GroupViT/datasets/imagenet\_template.py}. Additionally, each batch includes only unique classes and the batch size is equal to the total number of classes. This way we prevent images and texts belonging to the same class from being pushed apart during contrastive learning. Few-shot learning methods are then compared between \textbf{two settings}: (1) CLIP trained from scratch on the full ImageNet dataset after unlearning a subset, and (2) CLIP trained from scratch while excluding that subset from the training data. Our goal is to determine whether the performance of these models is comparable, indicating that unlearning can approximate the effects of training from scratch. % \subsection{Results of Unlearning and Oracle Baselines}

\begin{figure}[!t]
    \centering
    \includegraphics[width=0.93\linewidth]{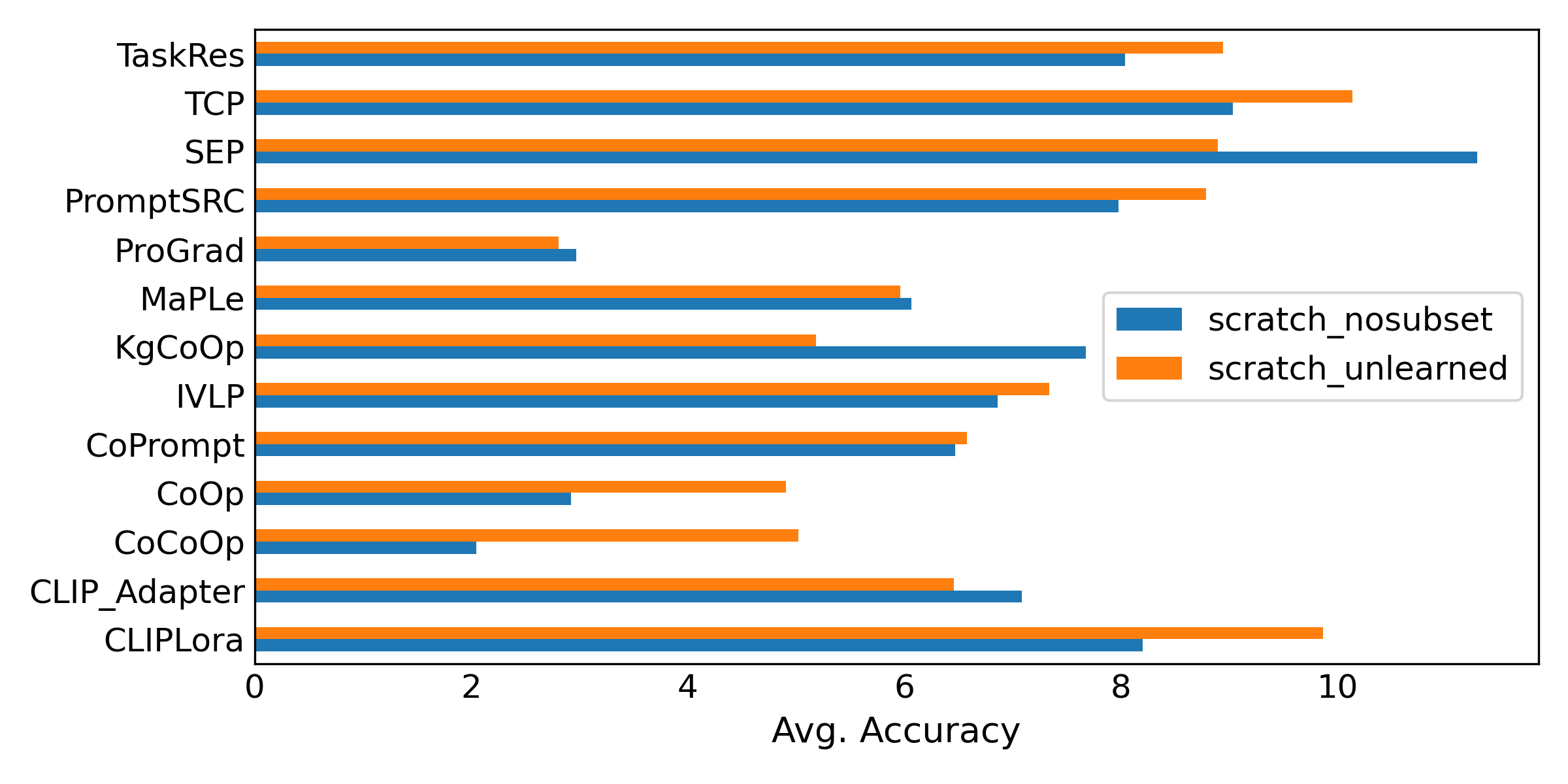}
    \caption{ Comparison of unlearned CLIP and CLIP trained excluding a subset from training in few-shot methods performance in the oracle setting. }
    \label{fig:scratch_methods_nosepres}
\vspace{-2mm}
\end{figure}

\paragraph{Results of Unlearning and Oracle Baselines}
Two settings are compared based on: (1) few-shot performance and (2) UMAP \cite{2018arXivUMAP} of visual features across the two settings. Tab.~\ref{tab:scrath_results} shows that the accuracy on the unlearned subset and on the remaining classes for both settings are similar which validates the previous evaluations. In Fig. \ref{fig:scratch_methods_nosepres} aggregated results across different few-shot methods are similar in both settings. Additionally, Fig. \ref{fig:umap_scratch_wlabels} shows the visual features projected into 2D space via UMAP across different settings for ''birds'' subset. UMAP with other subsets is in Supplementary (Section VI). \textit{No subset} represents CLIP trained from scratch excluding the subset, \textit{unlearned} is CLIP trained from scratch on full data and unlearning the subset, and \textit{full} is CLIP trained from scratch on full dataset. Fig.~\ref{fig:umap_scratch_wlabels} shows different colour samples from the subset that is excluded and samples belonging to other classes. Some classes from the excluded subset are highlighted. We observe that: (1) in both \textit{No subset} and \textit{unlearned}, the highlighted classes from the excluded subset are more sparse and overlapping compared to \textit{full} where samples from the excluded subset belonging to the same class are more clustered together and do not overlap with other classes (clear separation). (2) the unlearned subset in both \textit{No subset} and \textit{unlearned} overlap much more with other classes compared to \textit{full}. 

Hence, unlearning and training from scratch have a similar effect.  Thus, the inductive evaluation to assess the true generalization capabilities of the few-shot learning methods proposed in our pipeline is valid.

\paragraph{Performance of SEPRES in the Oracle Setting}
Tab. \ref{tab:scratch_subsets_avg} shows a comparison of few-shot learning methods in the oracle setting and granular results provided in the Supplementary (Section VII). SEPRES consistently outperforms other baseline methods when CLIP is trained from scratch excluding a subset, in line with unlearned setting. We also note that MMD scores between \textit{Dogs} subset and the remaining ImageNet dataset is higher compared to \textit{Birds} and \textit{Vehicles} as shown in Tab. \ref{tab:scrath_results} in the last column. Hence, there is %This explains 
a smaller accuracy increase on \textit{Dogs} dataset, as effectively there is less related information of ``dogs" in CLIP that SEPRES can use to improve CLIP's performance. % on this subset. 

\subsection{Benchmark Setting Analysis}
In this paper, all the components of our pipeline are analyzed. We experiment with:
(a) 4 different unlearning settings - in this setting we lose general knowledge of the CLIP model by less than 3/\%, 25/\%, 50/\% and 90/\% while unlearning a particular selected dataset; (b) 14 different few-shot classification methods, including the proposed SEPRES method; (c) 7 different forget datasets by using the corresponding set of retain datasets shown in Tab.~\ref{tab:forget_results} and validating on the 5 validation datasets; (d) 5 different few-shot settings - (1,2,4,8 and 16 shots) with 3 seeds for each setting.
The results are summarized in Fig.~\ref{fig:aggressive_unlearning} that shows the average accuracy across all 7 datasets for each method. In total we conduct \textbf{5880 experiments} for a thorough analysis. As more knowledge is lost, accuracy across all methods goes down, but the proposed SEPRES method tends to be more robust compared to other methods. Granular results for this analysis are in the Supplementary (Sections IX-XI) where we also provide ablations on SEPRES (Section III).

\section{Conclusions}

In this paper, we presented a general pipeline designed to create a benchmark for large-scale model based few-shot learning methods in an inductive manner, which is important especially when the training data of the model is unknown. We applied it to CLIP-based few-shot methods demonstrating that when benchmarking is performed in an inductive way using the proposed pipeline, the performance of the baselines substantially drops by 55\% on average. This reveals a gap in current few-shot learning methods' ability to generalize to novel classes which are unseen in the training set of large models. A CLIP-based few-shot method (SEPRES) is proposed which achieves state-of-the-art results in various settings. Although we focused on CLIP, the pipeline can be adapted to any model with an effective unlearning strategy. Future work will extend this approach to large language and vision-language models with architectures that are varied from CLIP.

\clearpage
\newpage
{   
    \paragraph{Acknowledgements}
    We'd like to gratefully acknowledge the support from the University of Bath for the studentship. 
    \small
    \bibliographystyle{ieeenat_fullname}
    \bibliography{main}
}

\renewcommand{\theenumi}{\Roman{enumi}}%

{
   \twocolumn[
   \Large
   \thetitle 
    \centering
    \vspace{0.5em}
    \\ Supplementary Material \\ 
    \vspace{1.5em}
    % \hfill 
    \raggedright 
    \textbf{Table of Contents} 
    \vspace{1em}
    \large
    \begin{enumerate}
    \item \textsc{Ease of Benchmarking and Compute Requirements } \hfill \pageref{app:compute}\\
    \item \textsc{Mathematical Formulation of SEPRES: referred to Sec. 4.3 main paper} \hfill \pageref{app:sepres_math}\\
    \item \textsc{SEPRES Ablations : referred to Sec. 5.4 main paper} \hfill \pageref{app:sep_abls}\\
    \item \textsc{Unlearning Analysis: referred to Sec. 4.1 main paper} \hfill \pageref{app:ablations_unl}\\
    \item \textsc{Discussion } \hfill \pageref{app:discussion}\\
    % \item \textsc{Limitations } \hfill \pageref{app:limitations}\\
    \item \textsc{UMAP Visual Features Distribution for all Subsets: referred to Sec. 5.3 main paper } \hfill \pageref{app:umap_all}\\
    \item \textsc{CLIP Trained from Scratch Results: referred to Sec. 5.3 main paper} \hfill \pageref{app:clip_scratch}\\
    \item \textsc{CLIP Unlearned Default Results: referred to Sec. 5.2 main paper} \hfill \pageref{app:clip_unl_default}\\
    \item \textsc{CLIP Aggressive Unlearning - 25\% Knowledge Lost Results: referred to Sec. 5.4 main paper} \hfill \pageref{app:clip_unl_25}\\
    \item \textsc{CLIP Aggressive Unlearning - 50\% Knowledge Lost Results: referred to Sec. 5.4 main paper } \hfill \pageref{app:clip_unl_50}\\
    \item \textsc{CLIP Aggressive Unlearning - 90\% Knowledge Lost Results: referred to Sec. 5.4 main paper } \hfill \pageref{app:clip_unl_90}\\
    \item \textsc{CLIP No Unlearning Results: referred to Sec. 5.2 main paper } \hfill \pageref{app:clip_nounl}\\
    \item \textsc{CLIP Aggressive Unlearning - Aggregated Results: referred to Sec. 5.4 main paper } \hfill \pageref{app:aggunl_aggregated}\\
    
    \end{enumerate}
    \vspace{3.0em}
   ] %< twocolumn
}

\section{Ease of Benchmarking and Compute Requirements}

\label{app:compute}

Our main experiments can be performed on a single NVIDIA GeForce RTX 3090 GPU with 24GB of memory. Tab. \ref{tab:proctime} displays the processing time required to compute the gradients for unlearning CLIP across various datasets. Hence, the process of creating a new benchmark is quick and allows rapid setup for additional datasets. Furthermore, our benchmarking pipeline is not only very rapid to setup, but also easy to test with novel few-shot learning methods. Once a CLIP model has been unlearned, it is ready to be tested on a new benchmark; testing it with a new few-shot method is straightforward and requires only a change in the checkpoint file. 

To train CLIP from scratch, we used 4 NVIDIA GeForce RTX 3090 GPUs to process the entire batch size of 1000, which is the number of classes of ImageNet.

\begin{table}[h]
\centering
\resizebox{0.48\textwidth}{!}{\begin{tabular}{|l|c|c|}
\hline
\rowcolor{gray!20} \textbf{Dataset}          & \textbf{Number of Data Points} & \textbf{Processing Time (min:sec)} \\ \hline
StanfordDogs              & 6144                          & 01:00                              \\ \hline
StanfordCars              & 6528                          & 01:25                              \\ \hline
Caltech101                & 4160                          & 00:35                              \\ \hline
OxfordFlowers             & 4096                          & 00:44                              \\ \hline
CUB                       & 3648                          & 00:39                              \\ \hline
UCF101                    & 7680                          & 01:05                              \\ \hline
FGVCAircraft              & 3392                          & 00:57                              \\ \hline
\end{tabular}}
\caption{Processing times for selected datasets.}
\label{tab:proctime}
\end{table}

\section{Mathematical Formulation of SEPRES}

\label{app:sepres_math}

This section formalizes the mathematical framework underlying our proposed SEPRES method, introduced in Section 4.3 of the main paper. Our work builds upon the Self-Enhanced Prompt Tuning (SEP) method, which refines visual prompts using a Token Fusion Module (TFM). While SEP is highly effective at extracting knowledge from CLIP, as the latent learnable prompts are passed through CLIP’s frozen internal layers, they are constrained by CLIP’s intermediate representations, which limits the model’s ability to acquire new and independent knowledge not seen during pre-training. To address this limitation, we introduce learnable residual parameters into the final textual embeddings of CLIP, inspired by TaskRes approach. This enables the model to acquire new, independent knowledge from few-shot examples rather than merely conforming to CLIP’s existing representations that are constrained by its internal operations. In this section, we first provide the mathematical formulation of SEP and subsequently extend it to demonstrate the modifications leading to SEPRES.

\subsection*{Standard CLIP Classification}

For zero-shot classification with CLIP over $K$ classes, each class label is embedded into a textual prompt (e.g., "A photo of a \{class\}") and processed through CLIP's text encoder to generate a weight matrix $\mathbf{W}^{\text{clip}} \in \mathbb{R}^{K \times d}$, where $d$ is the embedding dimension. Given a test image, it is encoded into an embedding $\mathbf{G}^{\text{clip}} \in \mathbb{R}^{d}$ by CLIP's image encoder. Classification logits are computed as:
\begin{equation}
    \text{logits} = \mathbf{G}^{\text{clip}} (\mathbf{W}^{\text{clip}})^T, \quad \text{logits} \in \mathbb{R}^{K}.
\end{equation}

\subsection*{SEPRES Method}

Let $\mathbf{E} \in \mathbb{R}^{n_e \times D}$ denote the image embeddings, where $n_e$ is the token length and $D$ is the hidden dimension. We initialize a learnable global visual prompt $\mathbf{Q} \in \mathbb{R}^{n_p \times D}$, where $n_p$ is the length of the prompt. These are concatenated as:
\begin{equation}
    \mathbf{Z}_0 = [\mathbf{E}, \mathbf{Q}] \in \mathbb{R}^{(n_e + n_p) \times D}.
\end{equation}
The sequence is passed through the first visual encoder layer $\psi_1$, resulting in:
\begin{equation}
    \mathbf{Z}_1 = \psi_1(\mathbf{Z}_0) \in \mathbb{R}^{(n_e + n_p) \times D}.
\end{equation}

The Token Fusion Module (TFM) refines the visual embeddings by integrating information from frozen and learned tokens. $\mathbf{Z}_1$ is split into pre-trained tokens $\mathbf{Z}_1^v$ (first $n_e$ tokens) and prompt tokens $\mathbf{Z}_1^p$ (last $n_p$ tokens). The TFM updates the prompt embeddings through cross-attention:
\begin{equation}
    \tilde{\mathbf{Z}}^p = \text{TFM}(\mathbf{Z}^v, \mathbf{Z}^p) = \text{softmax}\left(\frac{\tilde{\mathbf{Z}}^v (\mathbf{Z}^p)^T}{\sqrt{D}}\right) \tilde{\mathbf{Z}}^v,
\end{equation}
where $\tilde{\mathbf{Z}}^v$ denotes the top-$n_p$ tokens with the highest activation scores, calculated as the mean squared values of token features.

The final visual token output for layer 1 is:
\begin{equation}
    \tilde{\mathbf{Z}}_1 = [\mathbf{Z}_1^v, \tilde{\mathbf{Z}}_1^p],
\end{equation}
which is passed to subsequent encoder layers:
\begin{equation}
    \tilde{\mathbf{Z}}^{l+1} = \psi_{l+1}([\mathbf{Z}_l^v, \text{TFM}(\mathbf{Z}_l^v, \mathbf{Z}_l^p)]), \quad l \ge 1.
\end{equation}

Similarly, enhanced textual tokens $\tilde{\mathbf{T}}_l$ are computed.

\paragraph{Enhanced Classifier with Residual Parameters}

With the enhanced prompts, an updated classifier's weight matrix $\mathbf{W}^{\text{sep}}$ and visual prompt $\mathbf{G}^{\text{sep}}$ are generated. We introduce learnable residual parameters to $\mathbf{W}^{\text{sep}}$. This approach enables the model to acquire new, independent knowledge rather than merely conforming to CLIP’s existing representations that are constrained by its internal operations. The resulting classifier weight matrix is defined as:

\begin{equation}
    \mathbf{W}^{\text{sepres}} = \mathbf{W}^{\text{sep}} + \alpha \mathbf{Y},
\end{equation}
where $\mathbf{Y}$ are learnable residual textual parameters of the same shape as $\mathbf{W}^{\text{sep}}$, initialized to zeros, and $\alpha$ is a scaling constant.

\paragraph{Optimization}

During training, the global visual/textual prompts and residual textual parameters are optimized using the following loss function:
\begin{align}
L &= \mathcal{L}_{\text{ce}}(\mathbf{G}^{sep}, \mathbf{W}^{\text{sepres}}) + \omega_t \mathcal{L}_{\text{kg}}(\mathbf{W}^{\text{clip}}, \mathbf{W}^{\text{sepres}}) \\ 
&\quad + \omega_v \mathcal{L}_{\text{kg}}(\mathbf{G}^{sep}, \mathbf{G}^{clip}) + \mathcal{L}_{\text{ce}}(\mathbf{G}^{clip}),
\end{align}
where:
\begin{itemize}
    \item $\mathcal{L}_{\text{ce}} = - \sum_{i=1}^{N} y_i \log(\hat{y}_i)$: Cross-entropy loss for classification.
    \item $\mathcal{L}_{\text{kg}} = \left\lVert A - B \right\rVert_2^2$: Knowledge-guided loss between weight matrices.
    \item $\omega_t, \omega_v$: Balancing coefficients.
\end{itemize}

\section{SEPRES Ablations}
\label{app:sep_abls}

\begin{figure}[h]
\centering
\hspace{-15mm}
\centering
\includegraphics[width=0.35\textwidth]{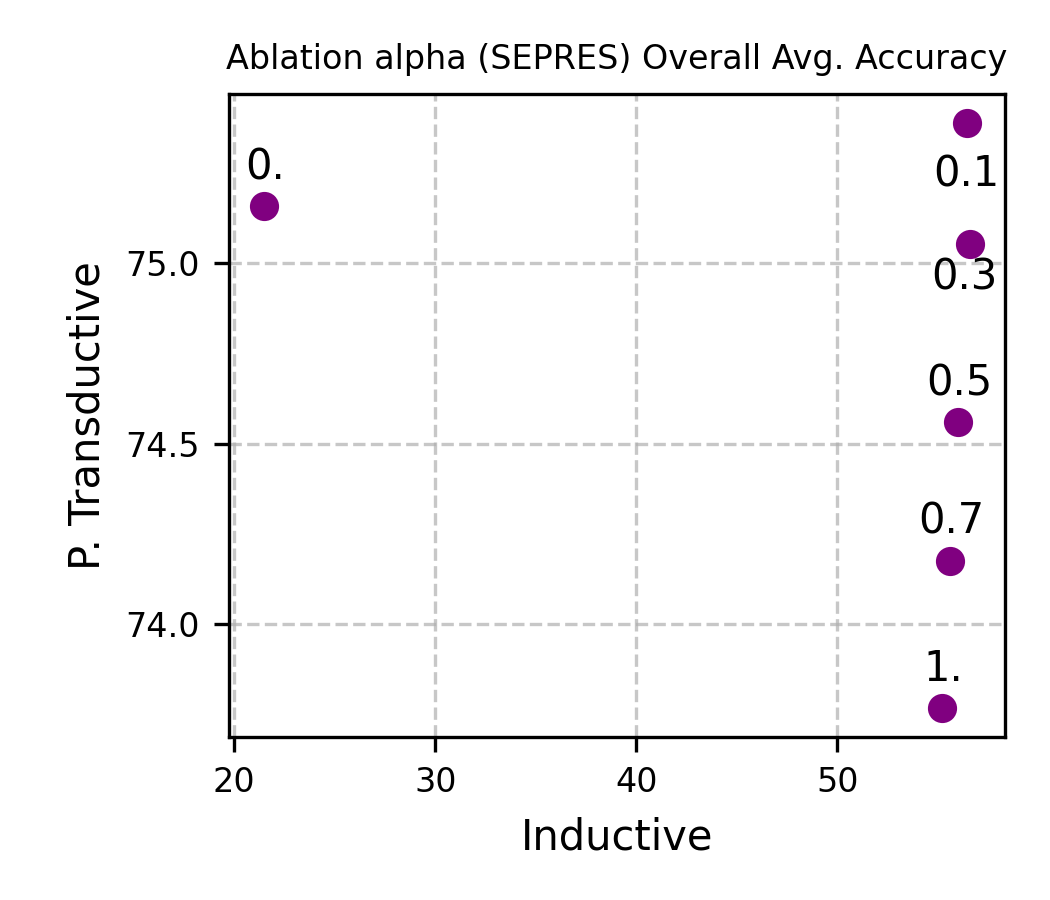} 
\caption{Ablation on $\alpha$ parameter of SEPRES. }

\label{fig:ablation_alpha}
\end{figure}

\subsection{Weight of Residual Features Ablation}
We conduct an ablation study for the residual feature importance in SEPRES, shown in Fig. \ref{fig:ablation_alpha}. Without residual features ($\alpha = 0.$) the accuracy is worse in both settings but mostly in P. Transductive, where the residual features are essential to learn new, independent knowledge to generalize in less known domains, which were removed through unlearning. Increasing $\alpha$ from 0.1 towards 1, the accuracy declines in the P. Transductive setting and slightly in the inductive one. We achieve the best result for $\alpha=0.1$.

\subsection{Visual Features and Components Ablation}

In Tab.~\ref{tab:ablation} we analyze the impact of adding the residual component on visual features (vRES). Incorporating vRES results in marginally lower accuracy while adding extra parameters to the model. The ablation study further demonstrates that both SEP and RES components are essential to the SEPRES method, as their individual removal leads to a performance drop. 

\begin{table}[!h]
    \centering
    \resizebox{0.35\textwidth}{!}{ % Changed 0.4 to ! to maintain aspect ratio
    \begin{tabular}{|l|c|c|}
    \hline
    \rowcolor{gray!20}
    \textbf{Ablation} & \textbf{Acc. Inductive} & \textbf{Acc. P. Transductive} \\
    \hline
    only SEP & 11.511 & 65.271 \\
    only RES & 21.492 & 75.157 \\
    SEPRES & \textbf{56.428} & \textbf{75.387} \\
    SEPRES + vRES & 56.17 & 75.021 \\
    \hline
    \end{tabular}}
    \caption{Ablation study on SEPRES components (SEP and RES) and on visual RES (vRES).}
    \label{tab:ablation}
\end{table}

\section{Unlearning Analysis}
\label{app:ablations_unl}

In this section, we provide additional unlearning analysis extending Sec. 4.1 of the main paper. Specifically, we present additional unlearning techniques evaluated in our experiments. We explored two zero-shot unlearning methods, one utilizing Lipschitz regularization (Lip) \cite{clipkrav} and another based on adjusting textual projection weights (TextProj) \cite{clipkravtextproj}. Additionally, we experimented with the non-zero-shot unlearning approach, SalUN \cite{fan2023salun}. These are compared against the SSD method we utilized in our main experiments, described in Sec. 4.1 of the main paper.

\begin{table}[!h]
    \centering
        \resizebox{0.45\textwidth}{!}{ % Resize table to fit the minipage
        \begin{tabular}{|c|c|c|c|}
        \hline
        \textbf{Forget Ds} & \textbf{\makecell{Unlearning \\ Method}} & \textbf{\makecell{Avg. Forget \\ Ds Acc. ($\downarrow$)}} & \textbf{\makecell{Avg. Knowledge  \\ Lost ($\downarrow$)}} \\ \hline
        Dogs, Cars, Flowers & SSD (Our) & 0.024 & 0.0283 \\ \hline
        Dogs, Cars, Flowers  & SalUn \cite{fan2023salun} & 0.0 & 0.3617 \\ \hline
        Dogs, Cars, Flowers  & Lip \cite{clipkrav} & 0.1907 & 0.3053 \\ \hline
        Dogs, Cars, Flowers & TextProj \cite{clipkravtextproj} & 0.043 & 0.0967 \\ \hline
        \end{tabular}}

\caption{Comparison of 4 unlearning methods on StanfordDogs, StanfordCars and OxfordFlowers datasets. SSD achieves the best trade-off between effective unlearning of targeted knowledge while minimizing unintended knowledge loss. } \label{tab:unl_ablations}

\end{table}

Tab. \ref{tab:unl_ablations} shows the average unlearning performance on the StanfordDogs, StanfordCars and OxfordFlowers datasets. The SSD method demonstrates the most balanced performance, effectively unlearning the target classes while minimizing the loss of knowledge on the validation set. SalUn — a non-zero-shot approach — achieves strong unlearning for targeted classes but struggles to preserve knowledge of non-targeted ones. Among the zero-shot methods, Lip is better at retaining knowledge than SalUn of non-targeted classes but falls short in unlearning the targeted classes. The TextProj method performs unlearning comparable to SSD for the targeted classes but incurs greater knowledge loss on the validation set, highlighting its trade-off limitations relative to SSD.

\section{Discussion}

\label{app:discussion}

Unlearning is a broad concept that cannot be easily guaranteed, and there are multiple metrics available to evaluate it. SSD unlearning was evaluated using accuracy and Membership Inference Attacks (MIA) \citep{shokri_2017_membership}, which assesses whether a particular data point was part of the training dataset used to train a machine learning model. For both metrics SSD has shown good results. Even if we do not know the exact data CLIP was trained on, we can assume that it has seen similar images of the standard few-shot evaluation benchmark datasets whose training set can be used for unlearning that knowledge. As our pipeline is general, when better unlearning methods with more theoretical unlearning guarantees become available, these can be used to create more robust benchmarks using our pipeline and understand true generalization of large models. Our pipeline is general and can be applied to any model as long as there is a good unlearning method available for it.

\section{UMAP Visual Features Distribution for all Subsets}

\label{app:umap_all}

\begin{figure}[!h]
    \onecolumn
    \centering
    \includegraphics[width=1.\textwidth]{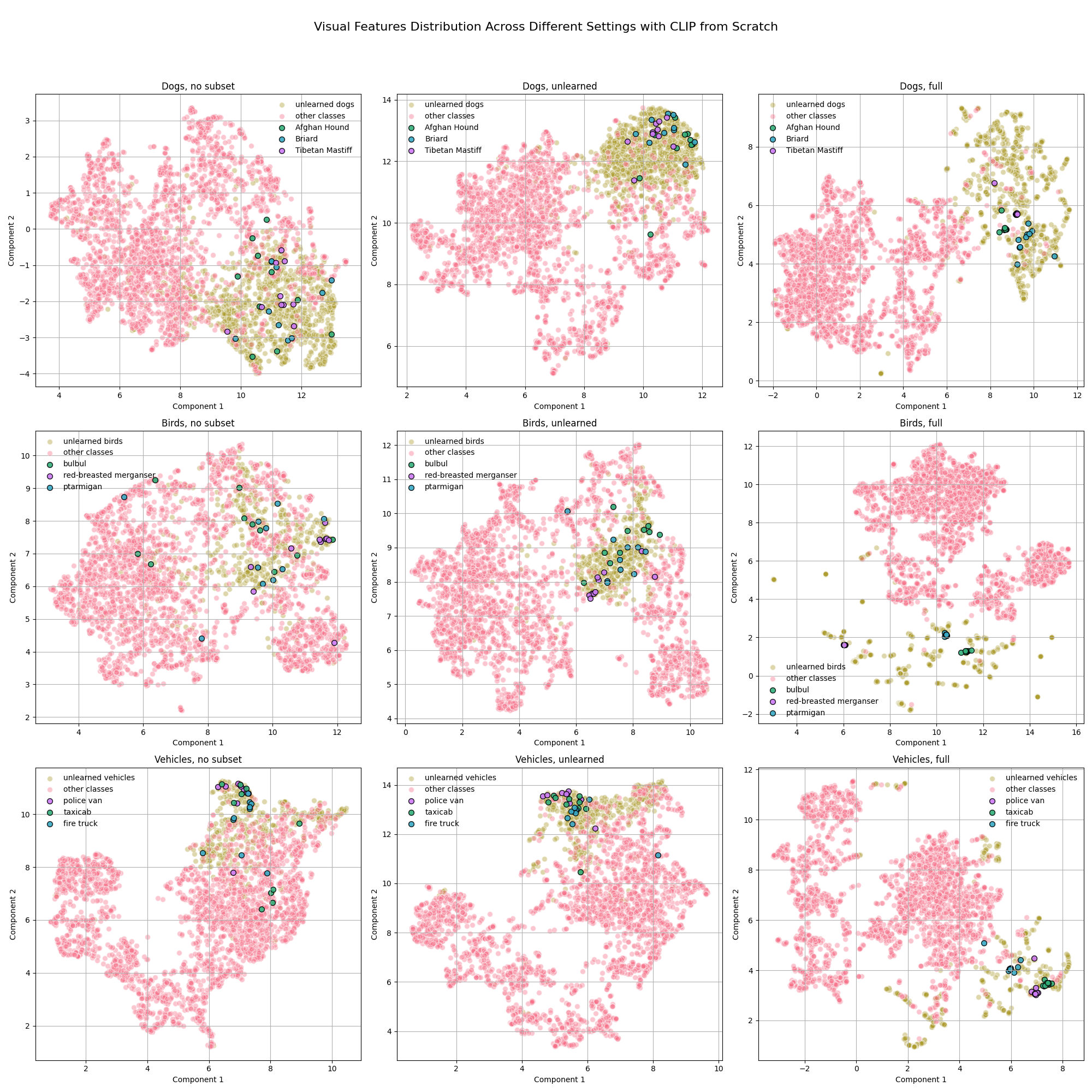}
   
    \caption{Visual features distributions across different settings with CLIP trained from scratch. Left visualisation: Excluding the subset from training. Middle visualisation: Unlearning the subset. Right visualisation: Without any unlearning. Classes to unlearn are shown in red while other classes are in light green. \textit{No subset} and \textit{unlearned} settings with the highlighted classes from the excluded subset are more sparse and overlapping compared to the \textit{full} setting. The unlearned subset in both \textit{no subset} and \textit{unlearned} settings overlap much more with other classes compared to full setting. All indicate that \textit{no subset} and \textit{unlearned} settings are similar. }
    \label{fig:umap_scratch_wlabels_all}
\end{figure}

In this section, we provide UMAP visual features visualizations on additional subsets that we discussed in Sec. 5.3. Fig. \ref{fig:umap_scratch_wlabels_all} shows these results for "birds", "vehicles" and "dogs" subsets from ImageNet. \textit{No subset} represents CLIP trained from scratch excluding the subset, \textit{unlearned} is CLIP trained from scratch on full data and unlearning the subset, and \textit{full} is CLIP trained from scratch on full dataset. In the figure, we used different colours for samples from the subset that is excluded and samples belonging to other classes. Some classes from the excluded subset are highlighted. As in our main paper that described the results for "birds", similar observations can be drawn for the other two subsets, namely that: (1) in both \textit{No subset} and \textit{unlearned}, the highlighted classes from the excluded subset are more sparse and overlapping compared to \textit{full} where samples from the excluded subset belonging to the same class are more clustered together and do not overlap with other classes (clear separation). (2) the unlearned subset in both \textit{No subset} and \textit{unlearned} overlap much more with other classes compared to \textit{full}. 

\clearpage
\twocolumn

\section{CLIP Trained from Scratch Full Results}
\label{app:clip_scratch}

In this section we present the full results for our oracle method discussed in Sec. 5.3 evaluated under two distinct settings - excluding a subset and unlearning a subset. Tab. \ref{tab:full_few_shot_results_excl} reports the performance of various few-shot learning methods with CLIP trained from scratch on the full ImageNet dataset, excluding the subset specified in the \textit{Dataset} column. The results are shown for different shot counts and subsets, averaged over three random seeds. Similarly, we present results for the unlearning setting in Tab. \ref{tab:full_few_shot_results_unl}.

Comparing the average performance across the two tables for different methods, we observe that the results are similar, demonstrating that excluding and unlearning the subset yield comparable outcomes. Furthermore, SEPRES achieves the best results in both scenarios, highlighting its effectiveness.

\begin{table}[h!]
\onecolumn
\centering
\resizebox{1.\textwidth}{!}{\begin{tabular}{|c|c||c|c|c|c|c|c|c|c|c|c|c|c|c|c|}
\hline
\rowcolor{gray!20} \textbf{Dataset} & \textbf{Shots} & \textbf{CLIPLora} & \textbf{CLIPAdapter} & \textbf{CoCoOp} & \textbf{CoOp} & \textbf{CoPrompt} & \textbf{IVLP} & \textbf{KgCoOp} & \textbf{MaPLe} & \textbf{ProGrad} & \textbf{PromptSRC} & \textbf{SEP} & \textbf{TCP} & \textbf{TaskRes} & \textbf{SEPRES} \\ \hline
Birds & 1 & 7.243 & 8.100 & 1.733 & 1.467 & 6.633 & 6.650 & 8.137 & 6.633 & 2.497 & 6.967 & 8.600 & 8.100 & 8.200 & \textbf{10.033} \\ \hline
Birds & 2 & 6.237 & 7.800 & 1.767 & 1.833 & 6.833 & 7.200 & 8.137 & 5.900 & 2.523 & 7.600 & 9.567 & 8.133 & 8.200 & \textbf{12.300} \\ \hline
Birds & 4 & 7.672 & 7.433 & 1.300 & 1.933 & 7.267 & 7.300 & 8.140 & 5.633 & 2.733 & 8.800 & 11.833 & 8.700 & 8.200 & \textbf{15.900} \\ \hline
Birds & 8 & 10.565 & 7.367 & 1.467 & 1.600 & 7.367 & 8.500 & 8.043 & 6.000 & 2.653 & 9.200 & 15.200 & 9.900 & 8.333 & \textbf{20.733} \\ \hline
Birds & 16 & 17.051 & 8.633 & 4.700 & 2.667 & 8.467 & 9.000 & 8.180 & 6.867 & 2.903 & 13.250 & 21.367 & 13.233 & 8.733 & \textbf{28.567} \\ \hline
Dogs & 1 & 3.928 & 5.967 & 1.200 & 1.900 & 5.967 & 4.567 & 6.467 & 6.367 & 2.463 & 5.100 & 5.667 & 5.667 & \textbf{6.567} & 6.400 \\ \hline
Dogs & 2 & 3.528 & 5.000 & 1.400 & 2.100 & 6.033 & 5.500 & 6.467 & 5.933 & 2.627 & 6.067 & 6.333 & 5.767 & 6.600 & \textbf{7.167} \\ \hline
Dogs & 4 & 3.983 & 5.467 & 2.100 & 2.033 & 5.900 & 5.333 & 6.467 & 5.233 & 2.533 & 5.867 & 8.067 & 6.233 & 6.633 & \textbf{9.200} \\ \hline
Dogs & 8 & 5.694 & 5.867 & 1.800 & 2.433 & 6.067 & 5.367 & 6.547 & 5.233 & 3.073 & 5.667 & 10.067 & 7.000 & 6.700 & \textbf{12.067} \\ \hline
Dogs & 16 & 7.578 & 6.333 & 2.000 & 4.500 & 6.267 & 6.500 & 6.450 & 5.000 & 3.457 & 5.233 & 11.300 & 7.767 & 7.067 & \textbf{13.467} \\ \hline
Vehicles & 1 & 6.389 & 7.733 & 1.967 & 3.400 & 8.367 & 6.067 & 8.467 & 8.233 & 3.237 & 6.700 & 7.867 & 9.067 & 8.467 & \textbf{12.100} \\ \hline
Vehicles & 2 & 7.344 & 7.867 & 1.867 & 3.467 & 7.667 & 6.943 & 8.353 & 7.600 & 2.633 & 8.633 & 9.367 & 9.167 & 8.500 & \textbf{14.067} \\ \hline
Vehicles & 4 & 8.389 & 7.233 & 1.967 & 3.800 & 5.067 & 6.067 & 8.400 & 3.967 & 3.600 & 8.667 & 11.267 & 10.867 & 8.567 & \textbf{18.367} \\ \hline
Vehicles & 8 & 11.233 & 7.067 & 2.700 & 5.067 & 4.667 & 8.433 & 8.387 & 6.600 & 3.980 & 10.433 & 13.567 & 12.233 & 9.267 & \textbf{24.633} \\ \hline
Vehicles & 16 & 16.144 & 8.433 & 2.767 & 5.633 & 4.467 & 9.467 & 8.467 & 5.800 & 3.613 & 11.433 & 19.300 & 13.633 & 10.533 & \textbf{30.100} \\ \hline
\hline
{Average} & 1 & 5.853 & 7.267 & 1.633 & 2.256 & 6.989 & 5.761 & 7.690 & 7.078 & 2.732 & 6.256 & 7.378 & 7.611 & 7.744 & \textbf{9.511} \\ \hline
{Average} & 2 & 5.703 & 6.889 & 1.678 & 2.467 & 6.844 & 6.548 & 7.652 & 6.478 & 2.594 & 7.433 & 8.422 & 7.689 & 7.767 & \textbf{11.178} \\ \hline
{Average} & 4 & 6.682 & 6.711 & 1.789 & 2.589 & 6.078 & 6.233 & 7.669 & 4.944 & 2.956 & 7.778 & 10.389 & 8.600 & 7.800 & \textbf{14.489} \\ \hline
{Average} & 8 & 9.164 & 6.767 & 1.989 & 3.033 & 6.033 & 7.433 & 7.659 & 5.944 & 3.236 & 8.433 & 12.944 & 9.711 & 8.100 & \textbf{19.144} \\ \hline
{Average} & 16 & 13.591 & 7.800 & 3.156 & 4.267 & 6.400 & 8.322 & 7.699 & 5.889 & 3.324 & 9.972 & 17.322 & 11.544 & 8.778 & \textbf{24.044} \\ \hline
Average (Overall) & - & 8.199 & 7.087 & 2.049 & 2.922 & 6.469 & 6.860 & 7.674 & 6.067 & 2.968 & 7.974 & 11.291 & 9.031 & 8.038 & \textbf{15.673} \\ \hline
\end{tabular}}
\caption{Few-shot results with CLIP trained from scratch on full ImageNet \textbf{excluding} the subset.}
\label{tab:full_few_shot_results_excl}
\end{table}

\begin{table}[h!]
\centering
\resizebox{1.\textwidth}{!}{\begin{tabular}{|c|c||c|c|c|c|c|c|c|c|c|c|c|c|c|c|}
\hline
\rowcolor{gray!20} \textbf{Dataset} & \textbf{Shots} & \textbf{CLIPLora} & \textbf{CLIPAdapter} & \textbf{CoCoOp} & \textbf{CoOp} & \textbf{CoPrompt} & \textbf{IVLP} & \textbf{KgCoOp} & \textbf{MaPLe} & \textbf{ProGrad} & \textbf{PromptSRC} & \textbf{SEP} & \textbf{TCP} & \textbf{TaskRes} & \textbf{SEPRES} \\ \hline
Birds & 1 & 8.147 & 5.867 & 2.133 & 3.567 & 5.900 & 6.900 & 4.907 & 5.800 & 2.623 & 7.633 & 6.400 & 7.300 & 6.200 & \textbf{15.933} \\ \hline
Birds & 2 & 9.729 & 6.933 & 5.300 & 4.367 & 6.533 & 7.367 & 4.907 & 6.100 & 2.420 & 8.600 & 7.933 & 8.433 & 6.400 & \textbf{21.300} \\ \hline
Birds & 4 & 11.989 & 8.267 & 5.567 & 4.067 & 8.133 & 9.033 & 5.017 & 7.267 & 3.357 & 11.267 & 11.800 & 13.200 & 6.567 & \textbf{32.833} \\ \hline
Birds & 8 & 15.932 & 9.367 & 8.167 & 4.967 & 9.600 & 10.633 & 5.093 & 7.933 & 3.480 & 14.067 & 10.533 & 16.633 & 9.633 & \textbf{47.067} \\ \hline
Birds & 16 & 22.983 & 10.667 & 9.467 & 6.400 & 10.567 & 13.300 & 5.713 & 8.533 & 3.267 & 17.200 & 24.967 & 22.267 & 16.300 & \textbf{56.433} \\ \hline
Dogs & 1 & 4.778 & 4.567 & 2.900 & 4.033 & 4.333 & 4.000 & 4.340 & 4.233 & 1.287 & 4.467 & 4.800 & 5.567 & 4.967 & \textbf{7.533} \\ \hline
Dogs & 2 & 5.572 & 4.800 & 4.033 & 4.033 & 4.633 & 4.233 & 4.357 & 4.267 & 0.903 & 5.667 & 5.133 & 6.400 & 5.067 & \textbf{10.633} \\ \hline
Dogs & 4 & 6.389 & 4.767 & 4.400 & 4.767 & 4.900 & 5.733 & 4.507 & 4.433 & 3.500 & 6.667 & 5.533 & 7.267 & 5.967 & \textbf{16.233} \\ \hline
Dogs & 8 & 7.922 & 4.867 & 4.700 & 4.733 & 5.367 & 6.400 & 4.497 & 5.133 & 4.413 & 7.067 & 5.967 & 9.067 & 9.500 & \textbf{23.000} \\ \hline
Dogs & 16 & 9.911 & 5.100 & 4.933 & 4.633 & 6.167 & 7.300 & 4.453 & 5.300 & 4.020 & 7.600 & 11.000 & 10.667 & 17.767 & \textbf{27.367} \\ \hline
Vehicles & 1 & 5.833 & 5.333 & 4.433 & 4.867 & 5.933 & 5.267 & 5.937 & 5.833 & 2.200 & 6.400 & 6.367 & 6.567 & 6.867 & \textbf{9.967} \\ \hline
Vehicles & 2 & 6.433 & 5.567 & 3.733 & 4.900 & 5.500 & 5.233 & 5.877 & 5.800 & 2.023 & 6.400 & 6.467 & 7.700 & 7.333 & \textbf{13.600} \\ \hline
Vehicles & 4 & 7.944 & 6.167 & 4.367 & 4.900 & 6.000 & 6.567 & 5.970 & 5.867 & 2.310 & 8.167 & 7.967 & 8.733 & 7.567 & \textbf{21.467} \\ \hline
Vehicles & 8 & 11.000 & 6.700 & 4.567 & 6.533 & 6.900 & 8.200 & 6.067 & 5.967 & 2.653 & 9.833 & 8.533 & 9.800 & 8.933 & \textbf{34.367} \\ \hline
Vehicles & 16 & 13.422 & 7.900 & 6.600 & 6.833 & 8.167 & 9.933 & 6.167 & 7.000 & 3.667 & 10.767 & 10.033 & 12.433 & 15.033 & \textbf{41.600} \\ \hline
\hline
{Average} & 1 & 6.253 & 5.256 & 3.156 & 4.156 & 5.389 & 5.389 & 5.061 & 5.289 & 2.037 & 6.167 & 5.856 & 6.478 & 6.011 & \textbf{11.144} \\ \hline
{Average} & 2 & 7.245 & 5.767 & 4.356 & 4.433 & 5.556 & 5.611 & 5.047 & 5.389 & 1.782 & 6.889 & 6.511 & 7.511 & 6.267 & \textbf{15.178} \\ \hline
{Average} & 4 & 8.774 & 6.400 & 4.778 & 4.578 & 6.344 & 7.111 & 5.164 & 5.856 & 3.056 & 8.700 & 8.433 & 9.733 & 6.700 & \textbf{23.511} \\ \hline
{Average} & 8 & 11.618 & 6.978 & 5.811 & 5.411 & 7.289 & 8.411 & 5.219 & 6.344 & 3.516 & 10.322 & 8.344 & 11.833 & 9.356 & \textbf{34.811} \\ \hline
{Average} & 16 & 15.439 & 7.889 & 7.000 & 5.956 & 8.300 & 10.178 & 5.444 & 6.944 & 3.651 & 11.856 & 15.333 & 15.122 & 16.367 & \textbf{41.800} \\ \hline
Average (Overall) & - & 9.866 & 6.458 & 5.020 & 4.907 & 6.576 & 7.340 & 5.187 & 5.964 & 2.808 & 8.787 & 8.896 & 10.136 & 8.940 & \textbf{25.289} \\ \hline
\end{tabular}}
\caption{Few-shot results with CLIP trained from scratch on full ImageNet \textbf{unlearning} the subset.}
\label{tab:full_few_shot_results_unl}
\end{table}

\twocolumn
\clearpage
\section{Original CLIP Full Results }

\subsection{Default runs unlearned CLIP.}
\label{app:clip_unl_default}

In this section, we present full results across different methods, number of shots and datasets after unlearning CLIP. Aggregated results are discussed in Sec. 5.2 of the main paper. SEPRES consistently outperforms other methods across all datasets and shot configurations. For the \textit{FgvcAircraft} dataset, the performance of SEP and CLIP-LoRA is comparable to SEPRES. However, these methods fall short across other datasets, demonstrating their lack of consistency. In contrast, SEPRES reliably delivers superior performance across all settings, highlighting its robustness and effectiveness.

\begin{table}[h!]
\onecolumn
\centering
\resizebox{1.\textwidth}{!}{\begin{tabular}{|c|c||c|c|c|c|c|c|c|c|c|c|c|c|c|c|}
\hline
\rowcolor{gray!20} \textbf{Dataset} & \textbf{Shots} & \textbf{CLIPLora} & \textbf{CLIPAdapter} & \textbf{CoCoOp} & \textbf{CoOp} & \textbf{CoPrompt} & \textbf{IVLP} & \textbf{KgCoOp} & \textbf{MaPLe} & \textbf{ProGrad} & \textbf{PromptSRC} & \textbf{SEP} & \textbf{TCP} & \textbf{TaskRes} & \textbf{SEPRES} \\ \hline
Stanforddogs & 0 & 0.020 & 0.020 & 0.020 & 0.020 & 0.020 & 0.020 & 0.020 & 0.020 & 0.020 & 0.020 & 0.020 & 0.020 & 0.020 & 0.020 \\ \hline
Stanforddogs & 1 & 3.515 & 2.000 & 2.900 & 3.467 & 2.733 & 2.267 & 3.120 & 1.233 & 3.287 & 5.533 & 5.733 & 4.850 & 2.200 & \textbf{14.767} \\ \hline
Stanforddogs & 2 & 7.226 & 2.167 & 2.500 & 6.133 & 3.000 & 6.200 & 3.447 & 1.667 & 3.830 & 10.100 & 10.300 & 8.367 & 2.333 & \textbf{27.367} \\ \hline
Stanforddogs & 4 & 11.592 & 2.467 & 3.200 & 9.733 & 5.733 & 9.500 & 5.067 & 3.233 & 6.910 & 14.400 & 14.500 & 13.100 & 2.700 & \textbf{44.300} \\ \hline
Stanforddogs & 8 & 15.279 & 2.733 & 4.150 & 13.600 & 9.967 & 16.433 & 7.340 & 4.600 & 11.247 & 17.867 & 17.933 & 16.067 & 4.700 & \textbf{57.333} \\ \hline
Stanforddogs & 16 & 19.458 & 4.000 & 7.150 & 17.333 & 15.133 & 18.933 & 9.450 & 9.867 & 15.977 & 19.833 & 20.233 & 18.350 & 11.767 & \textbf{67.933} \\ \hline
Stanfordcars & 0 & 0.011 & 0.011 & 0.011 & 0.011 & 0.011 & 0.011 & 0.011 & 0.011 & 0.011 & 0.011 & 0.011 & 0.011 & 0.011 & 0.011 \\ \hline
Stanfordcars & 1 & 3.971 & 1.167 & 1.167 & 3.050 & 1.233 & 1.600 & 1.400 & 1.550 & 1.720 & 4.300 & 5.000 & 3.533 & 1.333 & \textbf{7.667} \\ \hline
Stanfordcars & 2 & 8.602 & 1.200 & 1.333 & 5.300 & 2.333 & 4.633 & 1.490 & 1.000 & 3.610 & 7.333 & 9.400 & 6.533 & 1.533 & \textbf{16.067} \\ \hline
Stanfordcars & 4 & 17.937 & 1.433 & 2.033 & 8.350 & 2.867 & 7.200 & 0.990 & 1.467 & 3.657 & 12.800 & 18.100 & 10.800 & 1.733 & \textbf{32.033} \\ \hline
Stanfordcars & 8 & 29.897 & 1.967 & 2.533 & 11.600 & 6.067 & 16.100 & 3.127 & 2.533 & 7.200 & 25.133 & 27.900 & 15.633 & 2.833 & \textbf{52.800} \\ \hline
Stanfordcars & 16 & 39.129 & 3.033 & 3.833 & 15.700 & 11.367 & 26.167 & 5.493 & 4.700 & 9.827 & 34.333 & 35.133 & 20.267 & 6.100 & \textbf{69.633} \\ \hline
Oxfordflowers & 0 & 0.054 & 0.054 & 0.054 & 0.054 & 0.054 & 0.054 & 0.054 & 0.054 & 0.054 & 0.054 & 0.054 & 0.054 & 0.054 & 0.054 \\ \hline
Oxfordflowers & 1 & 20.395 & 6.133 & 7.100 & 23.633 & 6.400 & 7.833 & 7.103 & 5.833 & 7.130 & 31.300 & 31.067 & 24.167 & 5.767 & \textbf{62.000} \\ \hline
Oxfordflowers & 2 & 34.768 & 6.367 & 7.900 & 33.967 & 10.900 & 34.800 & 6.713 & 6.000 & 7.010 & 39.900 & 43.467 & 38.867 & 5.800 & \textbf{80.600} \\ \hline
Oxfordflowers & 4 & 42.360 & 6.800 & 10.400 & 40.567 & 19.367 & 41.867 & 8.607 & 13.233 & 12.083 & 45.200 & 47.000 & 45.533 & 6.133 & \textbf{90.300} \\ \hline
Oxfordflowers & 8 & 47.449 & 7.600 & 22.800 & 45.200 & 37.367 & 46.767 & 9.460 & 23.233 & 23.967 & 48.467 & 48.933 & 47.833 & 7.367 & \textbf{95.067} \\ \hline
Oxfordflowers & 16 & 50.020 & 10.867 & 32.500 & 48.033 & 45.667 & 49.433 & 12.910 & 41.933 & 38.760 & 50.000 & 49.967 & 49.000 & 15.900 & \textbf{97.567} \\ \hline
Caltech101 & 0 & 0.080 & 0.080 & 0.080 & 0.080 & 0.080 & 0.080 & 0.080 & 0.080 & 0.080 & 0.080 & 0.080 & 0.080 & 0.080 & 0.080 \\ \hline
Caltech101 & 1 & 12.319 & 8.000 & 7.767 & 10.367 & 10.567 & 13.533 & 9.073 & 7.033 & 8.707 & 13.833 & 14.767 & 13.000 & 13.867 & \textbf{71.567} \\ \hline
Caltech101 & 2 & 13.550 & 8.067 & 8.467 & 12.767 & 9.733 & 12.733 & 8.600 & 8.967 & 9.873 & 16.500 & 16.400 & 15.967 & 14.800 & \textbf{83.833} \\ \hline
Caltech101 & 4 & 16.619 & 8.133 & 9.533 & 15.133 & 13.867 & 13.767 & 11.020 & 11.833 & 13.930 & 17.500 & 17.367 & 16.933 & 22.067 & \textbf{89.700} \\ \hline
Caltech101 & 8 & 17.904 & 8.267 & 11.100 & 16.833 & 15.800 & 17.367 & 13.660 & 12.700 & 14.253 & 17.700 & 17.900 & 17.600 & 33.800 & \textbf{93.067} \\ \hline
Caltech101 & 16 & 18.039 & 8.533 & 12.767 & 17.300 & 15.800 & 18.033 & 16.770 & 16.267 & 15.120 & 18.167 & 18.200 & 17.767 & 59.833 & \textbf{94.767} \\ \hline
Fgvcaircraft & 0 & 0.009 & 0.009 & 0.009 & 0.009 & 0.009 & 0.009 & 0.009 & 0.009 & 0.009 & 0.009 & 0.009 & 0.009 & 0.009 & 0.009 \\ \hline
Fgvcaircraft & 1 & 3.530 & 1.100 & 1.267 & 4.133 & 1.467 & 2.933 & 2.120 & 1.467 & 2.230 & 5.400 & 6.167 & 4.867 & 1.000 & \textbf{8.833} \\ \hline
Fgvcaircraft & 2 & 9.671 & 1.000 & 1.767 & 6.467 & 2.400 & 4.367 & 1.840 & 1.800 & 2.990 & 9.000 & 12.967 & 9.067 & 0.967 & \textbf{13.533} \\ \hline
Fgvcaircraft & 4 & 17.902 & 1.333 & 2.067 & 11.700 & 2.567 & 9.433 & 4.400 & 1.967 & 5.400 & 16.767 & 21.633 & 15.600 & 1.167 & \textbf{23.600} \\ \hline
Fgvcaircraft & 8 & 31.613 & 2.367 & 3.800 & 15.133 & 6.300 & 16.700 & 5.600 & 3.600 & 8.650 & 27.500 & 32.967 & 21.700 & 1.700 & \textbf{34.300} \\ \hline
Fgvcaircraft & 16 & 46.115 & 3.633 & 4.533 & 20.167 & 9.900 & 31.800 & 10.400 & 6.067 & 13.760 & 41.667 & 46.033 & 28.733 & 3.333 & \textbf{47.967} \\ \hline
Ucf101 & 0 & 0.052 & 0.052 & 0.052 & 0.052 & 0.052 & 0.052 & 0.052 & 0.052 & 0.052 & 0.052 & 0.052 & 0.052 & 0.052 & 0.052 \\ \hline
Ucf101 & 1 & 8.195 & 5.400 & 4.733 & 6.800 & 7.467 & 7.333 & 7.727 & 3.533 & 8.203 & 9.867 & 10.367 & 8.300 & 6.833 & \textbf{37.933} \\ \hline
Ucf101 & 2 & 11.296 & 5.500 & 7.400 & 11.200 & 9.167 & 8.267 & 8.083 & 6.400 & 8.450 & 12.767 & 13.533 & 10.833 & 7.400 & \textbf{54.233} \\ \hline
Ucf101 & 4 & 13.490 & 6.000 & 8.067 & 13.333 & 10.800 & 13.233 & 10.213 & 8.700 & 11.437 & 14.667 & 15.300 & 13.900 & 9.733 & \textbf{65.633} \\ \hline
Ucf101 & 8 & 15.764 & 6.933 & 8.633 & 14.533 & 12.233 & 15.200 & 11.373 & 9.567 & 13.457 & 15.700 & 16.533 & 15.467 & 13.600 & \textbf{75.200} \\ \hline
Ucf101 & 16 & 17.076 & 7.967 & 9.367 & 16.167 & 14.733 & 16.533 & 13.817 & 12.433 & 14.513 & 16.967 & 17.667 & 16.867 & 28.367 & \textbf{80.833} \\ \hline
Cub & 0 & 0.015 & 0.015 & 0.015 & 0.015 & 0.015 & 0.015 & 0.015 & 0.015 & 0.015 & 0.015 & 0.015 & 0.015 & 0.015 & 0.015 \\ \hline
Cub & 1 & 3.625 & 2.233 & 2.333 & 5.067 & 2.367 & 3.000 & 1.533 & 1.733 & 2.603 & 8.133 & 7.900 & 5.967 & 1.600 & \textbf{25.867} \\ \hline
Cub & 2 & 7.721 & 2.167 & 2.767 & 8.000 & 3.450 & 8.350 & 2.003 & 1.833 & 5.887 & 12.767 & 13.867 & 11.500 & 1.867 & \textbf{44.667} \\ \hline
Cub & 4 & 13.376 & 2.167 & 4.633 & 11.133 & 4.250 & 13.600 & 2.563 & 3.367 & 9.117 & 16.967 & 19.733 & 16.600 & 2.567 & \textbf{61.567} \\ \hline
Cub & 8 & 19.537 & 2.533 & 5.733 & 16.533 & 10.100 & 17.900 & 4.123 & 4.467 & 11.817 & 21.567 & 23.333 & 21.467 & 5.900 & \textbf{73.033} \\ \hline
Cub & 16 & 23.063 & 3.900 & 8.300 & 19.233 & 12.400 & 20.900 & 7.430 & 10.700 & 14.513 & 23.600 & 24.933 & 23.667 & 21.067 & \textbf{79.400} \\ \hline \hline
Average & 0 & 0.035 & 0.035 & 0.035 & 0.035 & 0.035 & 0.035 & 0.035 & 0.035 & 0.035 & 0.035 & 0.035 & 0.035 & 0.035 & 0.035 \\ \hline
Average & 1 & 7.936 & 3.719 & 3.895 & 8.074 & 4.605 & 5.500 & 4.582 & 3.198 & 4.840 & 11.195 & 11.571 & 9.240 & 4.657 & \textbf{32.662} \\ \hline
Average & 2 & 13.262 & 3.781 & 4.590 & 11.976 & 5.855 & 11.336 & 4.597 & 3.952 & 5.950 & 15.481 & 17.133 & 14.448 & 4.957 & \textbf{45.757} \\ \hline
Average & 4 & 19.040 & 4.048 & 5.705 & 15.707 & 8.493 & 15.514 & 6.123 & 6.257 & 8.933 & 19.757 & 21.948 & 18.924 & 6.586 & \textbf{58.162} \\ \hline
Average & 8 & 25.349 & 4.629 & 8.393 & 19.062 & 13.976 & 20.924 & 7.812 & 8.671 & 12.941 & 24.848 & 26.500 & 22.252 & 9.986 & \textbf{68.686} \\ \hline
Average & 16 & 30.414 & 5.990 & 11.207 & 21.990 & 17.857 & 25.971 & 10.896 & 14.567 & 17.496 & 29.224 & 30.310 & 24.950 & 20.910 & \textbf{76.871} \\ \hline
Overall Average (excl. zs) & - & 19.200 & 4.433 & 6.758 & 15.362 & 10.157 & 15.849 & 6.802 & 7.329 & 10.032 & 20.101 & 21.492 & 17.963 & 9.419 & \textbf{56.428} \\ \hline
\end{tabular}}
\caption{Few-shot performance results for unlearned CLIP with \textit{minimum} level of knowledge lost evaluated across multiple datasets, methods, and shot counts. For shot counts greater than zero, the top-performing results are indicated in \textbf{bold}.  Average results are shown at the end of the table. The table also includes averages for each shot count, with a final overall average calculated by excluding the zero-shot results. }
\label{tab:default_few_shot_results}
\end{table}

\twocolumn

\clearpage
\subsection{Aggressive Unlearning with 25\% knowledge lost.}
\label{app:clip_unl_25}

In this section, we present full results across different methods, number of shots and datasets after unlearning CLIP losing 25\% of its general knowledge. Aggregated results are presented and discussed in Sec. 5.4 of the main paper. Even under this knowledge reduction, SEPRES consistently outperforms all other methods across datasets and shot configurations. 

When comparing the results to those in Tab. \ref{tab:default_few_shot_results}, SEPRES maintains robust performance, with the average accuracy dropping from 56.4\% to 48.8\% — a reduction of only 8\%. In contrast, the performance of other methods declines by approximately half, highlighting their vulnerability to general knowledge loss. This underscores SEPRES's resilience and effectiveness, even in scenarios with significant reductions in CLIP's knowledge.

\begin{table}[h!]
\onecolumn
\centering
\resizebox{1.\textwidth}{!}{\begin{tabular}{|c|c||c|c|c|c|c|c|c|c|c|c|c|c|c|c|}
\hline
\rowcolor{gray!20} \textbf{Dataset} & \textbf{Shots} & \textbf{CLIPLora} & \textbf{CLIPAdapter} & \textbf{CoCoOp} & \textbf{CoOp} & \textbf{CoPrompt} & \textbf{IVLP} & \textbf{KgCoOp} & \textbf{MaPLe} & \textbf{ProGrad} & \textbf{PromptSRC} & \textbf{SEP} & \textbf{TCP} & \textbf{TaskRes} & \textbf{SEPRES} \\ \hline
Stanforddogs & 0 & 0.007 & 0.007 & 0.007 & 0.007 & 0.007 & 0.007 & 0.007 & 0.007 & 0.007 & 0.007 & 0.007 & 0.007 & 0.007 & 0.007 \\ \hline
Stanforddogs & 1 & 1.228 & 0.700 & 0.967 & 1.733 & 1.133 & 1.267 & 0.993 & 0.967 & 1.330 & 1.733 & 1.967 & 1.633 & 2.600 & \textbf{9.633} \\ \hline
Stanforddogs & 2 & 2.067 & 0.667 & 1.367 & 2.167 & 1.567 & 1.567 & 0.783 & 1.100 & 1.363 & 2.767 & 3.100 & 2.500 & 3.033 & \textbf{17.367} \\ \hline
Stanforddogs & 4 & 2.715 & 0.667 & 1.167 & 2.700 & 1.333 & 1.867 & 0.997 & 1.200 & 2.003 & 3.200 & 3.833 & 3.267 & 6.300 & \textbf{31.367} \\ \hline
Stanforddogs & 8 & 4.343 & 0.700 & 1.667 & 3.767 & 1.767 & 2.700 & 1.637 & 1.833 & 3.203 & 4.300 & 4.867 & 4.100 & 11.433 & \textbf{47.567} \\ \hline
Stanforddogs & 16 & 5.134 & 0.833 & 1.967 & 4.633 & 3.200 & 4.900 & 3.347 & 2.300 & 4.520 & 4.933 & 5.567 & 4.867 & 18.600 & \textbf{60.300} \\ \hline
Stanfordcars & 0 & 0.006 & 0.006 & 0.006 & 0.006 & 0.006 & 0.006 & 0.006 & 0.006 & 0.006 & 0.006 & 0.006 & 0.006 & 0.006 & 0.006 \\ \hline
Stanfordcars & 1 & 2.139 & 0.867 & 0.733 & 2.067 & 1.367 & 1.467 & 1.117 & 0.667 & 1.320 & 2.400 & 2.467 & 1.900 & 0.600 & \textbf{6.100} \\ \hline
Stanfordcars & 2 & 4.659 & 0.900 & 1.067 & 3.800 & 1.233 & 1.767 & 1.497 & 0.733 & 3.133 & 4.733 & 5.900 & 3.433 & 0.600 & \textbf{12.700} \\ \hline
Stanfordcars & 4 & 10.592 & 1.067 & 1.100 & 5.967 & 1.867 & 4.500 & 1.843 & 1.433 & 5.020 & 8.033 & 11.500 & 6.533 & 0.700 & \textbf{24.767} \\ \hline
Stanfordcars & 8 & 20.661 & 1.400 & 1.933 & 8.333 & 3.400 & 8.767 & 2.720 & 2.000 & 7.687 & 15.900 & 17.600 & 10.000 & 1.200 & \textbf{43.433} \\ \hline
Stanfordcars & 16 & 30.552 & 1.867 & 2.567 & 10.867 & 6.067 & 17.833 & 2.753 & 3.300 & 9.543 & 24.467 & 25.033 & 13.300 & 6.267 & \textbf{61.467} \\ \hline
Oxfordflowers & 0 & 0.038 & 0.038 & 0.038 & 0.038 & 0.038 & 0.038 & 0.038 & 0.038 & 0.038 & 0.038 & 0.038 & 0.038 & 0.038 & 0.038 \\ \hline
Oxfordflowers & 1 & 10.137 & 3.400 & 4.133 & 11.067 & 3.767 & 2.400 & 4.305 & 3.033 & 5.280 & 14.567 & 15.667 & 11.400 & 4.400 & \textbf{48.300} \\ \hline
Oxfordflowers & 2 & 17.107 & 3.400 & 5.533 & 16.400 & 5.600 & 13.767 & 4.385 & 2.300 & 5.753 & 19.367 & 22.900 & 18.667 & 4.433 & \textbf{67.500} \\ \hline
Oxfordflowers & 4 & 23.603 & 3.833 & 9.600 & 21.033 & 8.933 & 21.733 & 5.990 & 6.533 & 7.457 & 24.967 & 26.267 & 23.200 & 5.233 & \textbf{81.500} \\ \hline
Oxfordflowers & 8 & 28.177 & 4.333 & 11.000 & 25.367 & 15.733 & 26.567 & 9.243 & 12.600 & 11.017 & 27.767 & 28.500 & 25.967 & 7.533 & \textbf{90.533} \\ \hline
Oxfordflowers & 16 & 31.222 & 6.133 & 13.967 & 27.267 & 23.133 & 28.600 & 12.993 & 18.067 & 17.730 & 29.433 & 29.533 & 27.567 & 16.367 & \textbf{95.367} \\ \hline
Caltech101 & 0 & 0.077 & 0.077 & 0.077 & 0.077 & 0.077 & 0.077 & 0.077 & 0.077 & 0.077 & 0.077 & 0.077 & 0.077 & 0.077 & 0.077 \\ \hline
Caltech101 & 1 & 9.790 & 7.233 & 5.433 & 7.733 & 6.800 & 9.667 & 8.277 & 6.233 & 8.303 & 10.400 & 10.533 & 8.633 & 25.933 & \textbf{60.133} \\ \hline
Caltech101 & 2 & 10.453 & 7.000 & 7.100 & 8.733 & 5.133 & 8.767 & 8.113 & 7.833 & 9.307 & 10.400 & 11.367 & 9.633 & 29.300 & \textbf{77.800} \\ \hline
Caltech101 & 4 & 11.778 & 6.400 & 6.833 & 10.600 & 8.533 & 11.733 & 8.317 & 9.700 & 10.413 & 11.400 & 12.333 & 11.033 & 41.567 & \textbf{86.967} \\ \hline
Caltech101 & 8 & 12.508 & 6.133 & 9.300 & 11.367 & 11.100 & 12.100 & 10.153 & 10.433 & 11.387 & 11.800 & 12.533 & 11.467 & 51.600 & \textbf{91.933} \\ \hline
Caltech101 & 16 & 12.603 & 6.500 & 9.900 & 12.267 & 10.767 & 12.433 & 9.857 & 11.533 & 12.130 & 12.400 & 12.767 & 12.233 & 67.400 & \textbf{94.400} \\ \hline
Fgvcaircraft & 0 & 0.011 & 0.011 & 0.011 & 0.011 & 0.011 & 0.011 & 0.011 & 0.011 & 0.011 & 0.011 & 0.011 & 0.011 & 0.011 & 0.011 \\ \hline
Fgvcaircraft & 1 & 3.070 & 1.067 & 1.700 & 3.400 & 1.767 & 2.833 & 1.730 & 1.933 & 2.350 & 5.367 & 5.767 & 4.267 & 1.100 & \textbf{8.033} \\ \hline
Fgvcaircraft & 2 & 7.651 & 1.133 & 2.167 & 5.833 & 2.367 & 3.800 & 1.370 & 1.567 & 2.110 & 7.867 & 11.467 & 7.400 & 1.133 & \textbf{11.967} \\ \hline
Fgvcaircraft & 4 & 15.972 & 1.300 & 2.300 & 10.067 & 3.000 & 6.600 & 2.810 & 2.133 & 6.040 & 15.133 & 20.100 & 12.767 & 1.400 & \textbf{20.267} \\ \hline
Fgvcaircraft & 8 & 28.393 & 1.633 & 4.067 & 13.800 & 4.800 & 16.067 & 3.390 & 3.567 & 10.860 & 25.400 & 30.933 & 18.133 & 1.767 & \textbf{30.967} \\ \hline
Fgvcaircraft & 16 & 44.304 & 2.467 & 4.833 & 18.400 & 8.933 & 26.433 & 5.450 & 5.433 & 15.720 & 38.733 & 43.667 & 25.267 & 3.200 & \textbf{44.667} \\ \hline
Ucf101 & 0 & 0.011 & 0.011 & 0.011 & 0.011 & 0.011 & 0.011 & 0.011 & 0.011 & 0.011 & 0.011 & 0.011 & 0.011 & 0.011 & 0.011 \\ \hline
Ucf101 & 1 & 1.727 & 1.033 & 1.200 & 2.000 & 1.800 & 1.167 & 1.893 & 1.300 & 1.667 & 3.533 & 3.633 & 2.467 & 7.667 & \textbf{28.833} \\ \hline
Ucf101 & 2 & 3.216 & 1.000 & 1.267 & 3.367 & 2.167 & 2.567 & 1.913 & 1.200 & 1.743 & 4.533 & 5.033 & 3.267 & 9.833 & \textbf{45.633} \\ \hline
Ucf101 & 4 & 4.661 & 1.100 & 2.000 & 4.167 & 3.000 & 3.633 & 1.973 & 1.833 & 3.487 & 5.133 & 5.467 & 4.500 & 14.700 & \textbf{58.267} \\ \hline
Ucf101 & 8 & 6.353 & 1.133 & 2.200 & 4.567 & 3.533 & 5.000 & 2.820 & 1.500 & 5.030 & 5.800 & 6.333 & 5.800 & 22.533 & \textbf{69.967} \\ \hline
Ucf101 & 16 & 6.705 & 1.333 & 3.500 & 5.833 & 5.267 & 6.367 & 3.500 & 2.867 & 5.673 & 6.667 & 6.733 & 6.300 & 36.900 & \textbf{76.233} \\ \hline
Cub & 0 & 0.008 & 0.008 & 0.008 & 0.008 & 0.008 & 0.008 & 0.008 & 0.008 & 0.008 & 0.008 & 0.008 & 0.008 & 0.008 & 0.008 \\ \hline
Cub & 1 & 1.688 & 1.400 & 1.100 & 1.233 & 0.800 & 1.400 & 1.400 & 1.167 & 1.340 & 1.867 & 1.633 & 1.900 & 1.900 & \textbf{12.400} \\ \hline
Cub & 2 & 2.066 & 1.467 & 1.333 & 1.800 & 1.300 & 2.450 & 1.597 & 1.300 & 2.110 & 2.650 & 2.533 & 2.633 & 2.267 & \textbf{25.300} \\ \hline
Cub & 4 & 2.949 & 1.400 & 1.333 & 2.267 & 1.400 & 3.150 & 1.850 & 1.833 & 2.607 & 2.700 & 4.033 & 2.967 & 3.900 & \textbf{42.150} \\ \hline
Cub & 8 & 4.124 & 1.400 & 1.867 & 3.200 & 2.267 & 3.500 & 1.953 & 1.967 & 3.243 & 3.500 & 4.967 & 3.733 & 9.533 & \textbf{57.650} \\ \hline
Cub & 16 & 5.905 & 1.800 & 2.200 & 3.700 & 3.133 & 4.950 & 2.100 & 2.500 & 3.983 & 5.350 & 5.900 & 4.933 & 18.667 & \textbf{68.750} \\ \hline \hline
Average & 0 & 0.022 & 0.022 & 0.022 & 0.022 & 0.022 & 0.022 & 0.022 & 0.022 & 0.022 & 0.022 & 0.022 & 0.022 & 0.022 & 0.022 \\ \hline
Average & 1 & 4.254 & 2.243 & 2.181 & 4.176 & 2.490 & 2.886 & 2.816 & 2.186 & 3.084 & 5.695 & 5.952 & 4.600 & 6.314 & \textbf{24.776} \\ \hline
Average & 2 & 6.746 & 2.224 & 2.833 & 6.014 & 2.767 & 4.955 & 2.808 & 2.290 & 3.646 & 7.474 & 8.900 & 6.790 & 7.229 & \textbf{36.895} \\ \hline
Average & 4 & 10.324 & 2.252 & 3.476 & 8.114 & 4.010 & 7.602 & 3.397 & 3.524 & 5.290 & 10.081 & 11.933 & 9.181 & 10.543 & \textbf{49.326} \\ \hline
Average & 8 & 14.937 & 2.390 & 4.576 & 10.057 & 6.086 & 10.671 & 4.560 & 4.843 & 7.490 & 13.495 & 15.105 & 11.314 & 15.086 & \textbf{61.721} \\ \hline
Average & 16 & 19.489 & 2.990 & 5.562 & 11.852 & 8.643 & 14.502 & 5.714 & 6.571 & 9.900 & 17.426 & 18.457 & 13.495 & 23.914 & \textbf{71.598} \\ \hline
Overall Average (excl. zs) & - & 11.150 & 2.420 & 3.726 & 8.043 & 4.799 & 8.123 & 3.859 & 3.883 & 5.882 & 10.834 & 12.070 & 9.076 & 12.617 & \textbf{48.863} \\ \hline
\end{tabular}}
\caption{Few-shot performance results for unlearned CLIP with \textit{25\%} of knowledge lost evaluated across multiple datasets, methods, and shot counts. For shot counts greater than zero, the top-performing results are indicated in \textbf{bold}.  Average results are shown at the end of the table. The table also includes averages for each shot count, with a final overall average calculated by excluding the zero-shot results.}
\label{tab:few_shot_results}
\end{table}

\clearpage
\twocolumn
\subsection{Aggressive Unlearning with 50\% knowledge lost.}
\label{app:clip_unl_50}

In this section, we present full results across different methods, number of shots and datasets after unlearning CLIP losing 50\% of its general knowledge. Aggregated results are presented and discussed in Sec. 5.4 of the main paper. Even under this knowledge reduction, SEPRES consistently outperforms all other methods across datasets and shot configurations.

\begin{table}[h!]
\onecolumn
\centering
\resizebox{1.\textwidth}{!}{\begin{tabular}{|c|c||c|c|c|c|c|c|c|c|c|c|c|c|c|c|}
\hline
\rowcolor{gray!20} \textbf{Dataset} & \textbf{Shots} & \textbf{CLIPLora} & \textbf{CLIPAdapter} & \textbf{CoCoOp} & \textbf{CoOp} & \textbf{CoPrompt} & \textbf{IVLP} & \textbf{KgCoOp} & \textbf{MaPLe} & \textbf{ProGrad} & \textbf{PromptSRC} & \textbf{SEP} & \textbf{TCP} & \textbf{TaskRes} & \textbf{SEPRES} \\ \hline
Stanforddogs & 0 & 0.009 & 0.009 & 0.009 & 0.009 & 0.009 & 0.009 & 0.009 & 0.009 & 0.009 & 0.009 & 0.009 & 0.009 & 0.009 & 0.009 \\ \hline
Stanforddogs & 1 & 1.338 & 0.933 & 1.167 & 1.100 & 0.933 & 1.267 & 1.137 & 0.833 & 1.013 & 1.467 & 1.600 & 1.267 & 2.733 & \textbf{7.300} \\ \hline
Stanforddogs & 2 & 1.607 & 1.000 & 1.067 & 1.733 & 1.133 & 1.433 & 1.207 & 0.933 & 1.310 & 2.600 & 2.533 & 2.000 & 3.133 & \textbf{13.033} \\ \hline
Stanforddogs & 4 & 2.627 & 1.067 & 1.400 & 2.500 & 1.233 & 1.733 & 1.350 & 1.100 & 1.767 & 2.733 & 3.767 & 2.833 & 5.167 & \textbf{25.633} \\ \hline
Stanforddogs & 8 & 3.659 & 1.067 & 1.567 & 3.467 & 1.600 & 2.933 & 2.140 & 1.533 & 2.817 & 3.600 & 4.733 & 3.733 & 8.433 & \textbf{39.333} \\ \hline
Stanforddogs & 16 & 4.589 & 1.200 & 2.133 & 4.000 & 2.233 & 4.333 & 2.773 & 1.567 & 4.003 & 4.367 & 5.100 & 4.300 & 14.167 & \textbf{53.150} \\ \hline
Stanfordcars & 0 & 0.006 & 0.006 & 0.006 & 0.006 & 0.006 & 0.006 & 0.006 & 0.006 & 0.006 & 0.006 & 0.006 & 0.006 & 0.006 & 0.006 \\ \hline
Stanfordcars & 1 & 1.393 & 0.567 & 0.700 & 1.533 & 0.800 & 1.633 & 0.720 & 0.500 & 1.400 & 1.533 & 2.200 & 1.533 & 0.600 & \textbf{4.600} \\ \hline
Stanfordcars & 2 & 2.682 & 0.600 & 0.867 & 2.300 & 1.400 & 1.667 & 0.880 & 1.033 & 2.430 & 2.800 & 2.500 & 2.050 & 0.733 & \textbf{7.600} \\ \hline
Stanfordcars & 4 & 5.609 & 0.667 & 1.133 & 3.767 & 1.300 & 1.967 & 1.070 & 1.200 & 3.860 & 4.500 & 6.700 & 2.900 & 0.933 & \textbf{15.400} \\ \hline
Stanfordcars & 8 & 12.063 & 0.967 & 1.367 & 5.067 & 2.000 & 5.267 & 1.397 & 1.033 & 5.180 & 9.400 & 9.950 & 5.150 & 2.333 & \textbf{27.800} \\ \hline
Stanfordcars & 16 & 23.065 & 1.433 & 1.800 & 7.333 & 3.167 & 9.967 & 1.713 & 2.567 & 7.637 & 14.700 & 10.750 & 7.400 & 6.300 & \textbf{44.900} \\ \hline
Oxfordflowers & 0 & 0.018 & 0.018 & 0.018 & 0.018 & 0.018 & 0.018 & 0.018 & 0.018 & 0.018 & 0.018 & 0.018 & 0.018 & 0.018 & 0.018 \\ \hline
Oxfordflowers & 1 & 10.908 & 2.667 & 3.300 & 9.200 & 2.700 & 5.433 & 2.760 & 1.767 & 5.075 & 12.667 & 15.533 & 11.200 & 3.700 & \textbf{41.367} \\ \hline
Oxfordflowers & 2 & 17.228 & 2.833 & 3.667 & 14.000 & 2.967 & 12.367 & 3.140 & 1.700 & 6.925 & 19.067 & 21.433 & 16.767 & 4.600 & \textbf{60.467} \\ \hline
Oxfordflowers & 4 & 23.021 & 2.867 & 6.800 & 18.867 & 4.567 & 20.033 & 4.870 & 4.133 & 10.980 & 23.700 & 25.067 & 21.767 & 6.300 & \textbf{75.733} \\ \hline
Oxfordflowers & 8 & 27.000 & 3.033 & 8.100 & 23.067 & 13.333 & 24.733 & 6.520 & 9.167 & 15.955 & 26.567 & 27.400 & 24.467 & 9.367 & \textbf{87.300} \\ \hline
Oxfordflowers & 16 & 30.072 & 4.633 & 10.400 & 24.933 & 19.633 & 28.500 & 11.287 & 16.333 & 22.330 & 28.633 & 28.833 & 26.333 & 16.633 & \textbf{93.300} \\ \hline
Caltech101 & 0 & 0.046 & 0.046 & 0.046 & 0.046 & 0.046 & 0.046 & 0.046 & 0.046 & 0.046 & 0.046 & 0.046 & 0.046 & 0.046 & 0.046 \\ \hline
Caltech101 & 1 & 5.882 & 4.733 & 7.333 & 7.700 & 6.667 & 8.733 & 5.680 & 6.833 & 8.540 & 9.767 & 8.600 & 7.000 & 9.800 & \textbf{48.550} \\ \hline
Caltech101 & 2 & 8.749 & 4.900 & 5.867 & 9.800 & 7.000 & 9.700 & 6.800 & 5.967 & 8.773 & 10.100 & 10.633 & 7.733 & 11.100 & \textbf{68.900} \\ \hline
Caltech101 & 4 & 10.196 & 5.033 & 6.767 & 10.300 & 8.500 & 10.533 & 6.547 & 7.133 & 6.670 & 10.567 & 11.200 & 7.500 & 16.467 & \textbf{79.950} \\ \hline
Caltech101 & 8 & 12.157 & 5.033 & 7.000 & 11.300 & 8.533 & 11.767 & 6.720 & 10.333 & 9.047 & 11.767 & 12.000 & 8.633 & 28.233 & \textbf{88.800} \\ \hline
Caltech101 & 16 & 12.306 & 5.133 & 7.167 & 11.250 & 11.900 & 11.767 & 8.963 & 11.000 & 10.360 & 12.200 & 12.567 & 9.667 & 51.300 & \textbf{92.200} \\ \hline
Fgvcaircraft & 0 & 0.011 & 0.011 & 0.011 & 0.011 & 0.011 & 0.011 & 0.011 & 0.011 & 0.011 & 0.011 & 0.011 & 0.011 & 0.011 & 0.011 \\ \hline
Fgvcaircraft & 1 & 2.840 & 0.833 & 1.467 & 3.300 & 1.700 & 2.600 & 1.060 & 1.333 & 3.560 & 5.067 & 5.667 & 3.267 & 1.100 & \textbf{6.667} \\ \hline
Fgvcaircraft & 2 & 6.961 & 0.900 & 1.733 & 5.367 & 2.167 & 4.733 & 1.380 & 1.867 & 3.890 & 7.367 & 10.167 & 5.533 & 1.067 & \textbf{10.900} \\ \hline
Fgvcaircraft & 4 & 14.721 & 1.100 & 2.000 & 9.400 & 3.300 & 8.367 & 1.440 & 1.833 & 7.730 & 12.633 & 18.900 & 9.833 & 1.000 & \textbf{19.067} \\ \hline
Fgvcaircraft & 8 & 26.663 & 1.633 & 3.233 & 12.700 & 4.500 & 14.533 & 1.720 & 3.200 & 11.590 & 24.533 & 30.567 & 14.800 & 1.300 & \textbf{30.733} \\ \hline
Fgvcaircraft & 16 & \textbf{43.684} & 2.500 & 4.333 & 16.033 & 6.967 & 28.267 & 4.220 & 6.100 & 15.190 & 36.533 & 41.400 & 20.733 & 3.400 & 43.333 \\ \hline
Ucf101 & 0 & 0.017 & 0.017 & 0.017 & 0.017 & 0.017 & 0.017 & 0.017 & 0.017 & 0.017 & 0.017 & 0.017 & 0.017 & 0.017 & 0.017 \\ \hline
Ucf101 & 1 & 1.692 & 1.567 & 1.300 & 1.367 & 1.400 & 1.267 & 0.943 & 1.300 & 1.490 & 2.500 & 2.533 & 1.450 & 3.667 & \textbf{19.467} \\ \hline
Ucf101 & 2 & 2.749 & 1.367 & 1.300 & 2.700 & 1.100 & 2.200 & 1.567 & 1.067 & 1.460 & 3.600 & 4.500 & 2.250 & 4.467 & \textbf{33.333} \\ \hline
Ucf101 & 4 & 3.551 & 1.200 & 1.100 & 2.500 & 1.967 & 2.967 & 1.490 & 1.400 & 2.793 & 4.667 & 4.500 & 2.567 & 7.467 & \textbf{46.233} \\ \hline
Ucf101 & 8 & 5.049 & 1.067 & 2.100 & 4.733 & 2.500 & 4.633 & 1.720 & 1.133 & 4.097 & 5.467 & 5.733 & 3.633 & 13.967 & \textbf{58.467} \\ \hline
Ucf101 & 16 & 5.763 & 1.500 & 2.500 & 5.067 & 4.067 & 5.333 & 2.803 & 1.867 & 5.040 & 5.833 & 6.367 & 5.367 & 23.967 & \textbf{67.867} \\ \hline
Cub & 0 & 0.008 & 0.008 & 0.008 & 0.008 & 0.008 & 0.008 & 0.008 & 0.008 & 0.008 & 0.008 & 0.008 & 0.008 & 0.008 & 0.008 \\ \hline
Cub & 1 & 1.104 & 1.067 & 0.800 & 1.067 & 1.300 & 1.367 & 1.013 & 0.900 & 1.277 & 1.333 & 1.150 & 1.000 & 1.800 & \textbf{9.367} \\ \hline
Cub & 2 & 1.417 & 1.100 & 1.100 & 1.433 & 1.200 & 1.533 & 0.990 & 1.100 & 1.573 & 1.933 & 1.800 & 1.350 & 2.500 & \textbf{19.433} \\ \hline
Cub & 4 & 2.179 & 1.100 & 1.300 & 1.900 & 1.333 & 2.167 & 0.940 & 1.133 & 2.030 & 2.367 & 3.400 & 1.650 & 3.200 & \textbf{34.900} \\ \hline
Cub & 8 & 3.732 & 1.100 & 1.467 & 2.567 & 1.350 & 3.100 & 1.120 & 1.300 & 2.410 & 2.700 & 4.400 & 2.050 & 7.750 & \textbf{52.067} \\ \hline
Cub & 16 & 5.321 & 1.133 & 1.567 & 3.133 & 1.900 & 4.133 & 1.313 & 1.900 & 2.920 & 3.333 & 5.533 & 2.600 & 14.200 & \textbf{63.200} \\ \hline \hline
Average & 0 & 0.016 & 0.016 & 0.016 & 0.016 & 0.016 & 0.016 & 0.016 & 0.016 & 0.016 & 0.016 & 0.016 & 0.016 & 0.016 & 0.016 \\ \hline
Average & 1 & 3.594 & 1.767 & 2.295 & 3.610 & 2.214 & 3.186 & 1.902 & 1.924 & 3.194 & 4.905 & 5.326 & 3.817 & 3.343 & \textbf{19.617} \\ \hline
Average & 2 & 5.913 & 1.814 & 2.229 & 5.333 & 2.424 & 4.805 & 2.280 & 1.952 & 3.766 & 6.781 & 7.652 & 5.383 & 3.943 & \textbf{30.524} \\ \hline
Average & 4 & 8.844 & 1.862 & 2.929 & 7.033 & 3.171 & 6.824 & 2.530 & 2.562 & 5.119 & 8.738 & 10.505 & 7.007 & 5.790 & \textbf{42.417} \\ \hline
Average & 8 & 12.903 & 1.986 & 3.548 & 8.986 & 4.831 & 9.567 & 3.048 & 3.957 & 7.299 & 12.005 & 13.540 & 8.924 & 10.198 & \textbf{54.929} \\ \hline
Average & 16 & 17.828 & 2.505 & 4.271 & 10.250 & 7.124 & 13.186 & 4.725 & 5.905 & 9.640 & 15.086 & 15.793 & 10.914 & 18.567 & \textbf{65.421} \\ \hline
Overall Average (excl. zs) & - & 9.816 & 1.987 & 3.054 & 7.042 & 3.953 & 7.513 & 2.897 & 3.260 & 5.803 & 9.503 & 10.563 & 7.209 & 8.368 & \textbf{42.581} \\ \hline
\end{tabular}}
\caption{Few-shot performance results for unlearned CLIP with \textit{50\%} level of knowledge lost evaluated across multiple datasets, methods, and shot counts. For shot counts greater than zero, the top-performing results are indicated in \textbf{bold}.  Average results are shown at the end of the table. The table also includes averages for each shot count, with a final overall average calculated by excluding the zero-shot results.}
\label{tab:few_shot_results}
\end{table}

\twocolumn
\clearpage
\subsection{Aggressive Unlearning with 90\% knowledge lost.}

\label{app:clip_unl_90}

In this section, we present full results across different methods, number of shots and datasets after unlearning CLIP losing 90\% of its general knowledge. This is almost equivalent of performing few-shot on an almost not trained CLIP model. Aggregated results are presented and discussed in Sec. 5.4 of the main paper. SEPRES still outperforms all other methods across most datasets and shot configurations, demonstrating SEPRES general applicability. 

\begin{table}[h!]
\onecolumn
\centering
\resizebox{1.\textwidth}{!}{\begin{tabular}{|c|c||c|c|c|c|c|c|c|c|c|c|c|c|c|c|}
\hline
\rowcolor{gray!20} \textbf{Dataset} & \textbf{Shots} & \textbf{CLIPLora} & \textbf{CLIPAdapter} & \textbf{CoCoOp} & \textbf{CoOp} & \textbf{CoPrompt} & \textbf{IVLP} & \textbf{KgCoOp} & \textbf{MaPLe} & \textbf{ProGrad} & \textbf{PromptSRC} & \textbf{SEP} & \textbf{TCP} & \textbf{TaskRes} & \textbf{SEPRES} \\ \hline
Stanforddogs & 0 & 0.008 & 0.008 & 0.008 & 0.008 & 0.008 & 0.008 & 0.008 & 0.008 & 0.008 & 0.008 & 0.008 & 0.008 & 0.008 & 0.008 \\ \hline
Stanforddogs & 1 & 0.880 & 0.900 & 0.767 & 0.800 & 0.767 & 0.800 & 0.850 & 0.700 & 0.773 & 0.733 & 0.767 & 0.700 & 1.967 & \textbf{2.200} \\ \hline
Stanforddogs & 2 & 0.868 & 0.900 & 0.833 & 0.800 & 0.833 & 0.867 & 0.813 & 0.833 & 0.827 & 0.867 & 0.933 & 0.800 & 2.467 & \textbf{2.633} \\ \hline
Stanforddogs & 4 & 1.076 & 0.900 & 0.767 & 0.767 & 0.700 & 0.867 & 0.820 & 0.767 & 0.733 & 0.733 & 0.900 & 0.767 & 2.800 & \textbf{3.467} \\ \hline
Stanforddogs & 8 & 1.160 & 0.967 & 0.867 & 0.800 & 0.800 & 0.867 & 0.853 & 0.933 & 0.750 & 0.833 & 1.233 & 0.867 & 3.200 & \textbf{5.267} \\ \hline
Stanforddogs & 16 & 1.639 & 0.967 & 0.933 & 0.800 & 0.933 & 1.033 & 0.807 & 0.967 & 0.913 & 0.900 & 1.167 & 0.933 & 4.000 & \textbf{9.067} \\ \hline
Stanfordcars & 0 & 0.006 & 0.006 & 0.006 & 0.006 & 0.006 & 0.006 & 0.006 & 0.006 & 0.006 & 0.006 & 0.006 & 0.006 & 0.006 & 0.006 \\ \hline
Stanfordcars & 1 & 1.003 & 0.633 & 0.600 & 0.900 & 0.600 & 0.767 & 0.680 & 0.667 & 0.803 & 0.833 & 1.000 & 0.700 & 1.233 & \textbf{2.267} \\ \hline
Stanfordcars & 2 & 1.356 & 0.600 & 0.767 & 1.267 & 0.867 & 0.733 & 0.820 & 0.767 & 1.187 & 1.067 & 1.300 & 0.900 & 1.233 & \textbf{3.200} \\ \hline
Stanfordcars & 4 & 2.114 & 0.600 & 0.733 & 1.700 & 0.867 & 1.267 & 0.860 & 0.867 & 1.650 & 1.200 & 1.433 & 1.067 & 1.267 & \textbf{4.867} \\ \hline
Stanfordcars & 8 & 4.469 & 0.600 & 1.033 & 2.267 & 0.967 & 1.833 & 0.817 & 0.833 & 2.220 & 1.833 & 0.700 & 1.367 & 2.000 & \textbf{7.867} \\ \hline
Stanfordcars & 16 & 8.258 & 0.700 & 1.200 & 3.000 & 0.967 & 3.200 & 1.100 & 1.100 & 2.897 & 2.967 & 1.200 & 2.000 & 3.233 & \textbf{13.650} \\ \hline
Oxfordflowers & 0 & 0.007 & 0.007 & 0.007 & 0.007 & 0.007 & 0.007 & 0.007 & 0.007 & 0.007 & 0.007 & 0.007 & 0.007 & 0.007 & 0.007 \\ \hline
Oxfordflowers & 1 & 6.171 & 0.700 & 0.967 & 3.367 & 1.433 & 2.600 & 0.703 & 1.567 & 2.573 & 4.633 & 4.967 & 2.600 & 3.333 & \textbf{16.833} \\ \hline
Oxfordflowers & 2 & 11.003 & 0.900 & 1.833 & 6.433 & 1.567 & 4.333 & 0.663 & 1.967 & 2.830 & 7.633 & 10.800 & 3.933 & 4.200 & \textbf{26.733} \\ \hline
Oxfordflowers & 4 & 15.970 & 1.233 & 1.767 & 10.967 & 2.267 & 8.633 & 1.513 & 2.367 & 8.283 & 11.767 & 17.067 & 7.933 & 7.700 & \textbf{38.433} \\ \hline
Oxfordflowers & 8 & 20.016 & 1.733 & 2.367 & 13.600 & 3.600 & 13.400 & 0.863 & 2.300 & 12.627 & 16.833 & 20.667 & 11.433 & 13.800 & \textbf{52.000} \\ \hline
Oxfordflowers & 16 & 23.549 & 2.267 & 3.033 & 16.000 & 5.100 & 18.167 & 2.857 & 3.267 & 15.767 & 20.633 & 23.867 & 14.767 & 25.200 & \textbf{68.233} \\ \hline
Caltech101 & 0 & 0.005 & 0.005 & 0.005 & 0.005 & 0.005 & 0.005 & 0.005 & 0.005 & 0.005 & 0.005 & 0.005 & 0.005 & 0.005 & 0.005 \\ \hline
Caltech101 & 1 & 0.933 & 0.500 & 1.700 & 1.167 & 1.433 & 1.233 & 0.823 & 0.600 & 0.977 & 0.900 & 0.900 & 0.933 & 1.767 & \textbf{2.433} \\ \hline
Caltech101 & 2 & 1.149 & 0.500 & 1.533 & 1.100 & 1.133 & 1.567 & 0.863 & 1.067 & 1.203 & 1.600 & 1.433 & 1.433 & 2.533 & \textbf{3.133} \\ \hline
Caltech101 & 4 & 1.596 & 0.500 & 1.233 & 1.100 & 1.067 & 1.033 & 0.957 & 1.000 & 1.107 & 1.300 & 1.200 & 1.000 & 2.767 & \textbf{6.200} \\ \hline
Caltech101 & 8 & 2.150 & 0.533 & 1.167 & 0.900 & 0.600 & 0.867 & 0.823 & 0.700 & 1.203 & 1.567 & 1.600 & 1.133 & 3.767 & \textbf{9.967} \\ \hline
Caltech101 & 16 & 2.637 & 0.567 & 1.200 & 1.167 & 0.800 & 1.267 & 1.070 & 1.233 & 1.393 & 1.600 & 2.167 & 1.567 & 5.133 & \textbf{10.700} \\ \hline
Fgvcaircraft & 0 & 0.012 & 0.012 & 0.012 & 0.012 & 0.012 & 0.012 & 0.012 & 0.012 & 0.012 & 0.012 & 0.012 & 0.012 & 0.012 & 0.012 \\ \hline
Fgvcaircraft & 1 & 1.720 & 1.267 & 1.133 & 1.133 & 1.033 & 1.533 & 1.140 & 1.233 & 1.040 & 1.467 & 1.433 & 1.333 & \textbf{1.933} & 1.800 \\ \hline
Fgvcaircraft & 2 & 1.580 & 1.267 & 1.067 & 1.100 & 1.233 & 1.700 & 1.040 & 1.367 & 1.310 & 1.567 & 1.733 & 1.200 & \textbf{2.533} & 2.367 \\ \hline
Fgvcaircraft & 4 & 2.120 & 1.300 & 1.600 & 1.167 & 1.633 & 1.833 & 1.560 & 1.333 & 1.050 & 1.900 & 1.767 & 1.267 & 2.967 & \textbf{3.300} \\ \hline
Fgvcaircraft & 8 & 3.150 & 1.300 & 1.700 & 1.200 & 1.833 & 1.867 & 1.710 & 1.633 & 1.030 & 1.900 & 2.033 & 1.433 & 2.900 & \textbf{4.000} \\ \hline
Fgvcaircraft & 16 & 3.650 & 1.267 & 1.800 & 1.700 & 1.800 & 1.867 & 1.610 & 1.833 & 1.050 & 1.800 & 2.300 & 1.867 & 3.533 & \textbf{4.933} \\ \hline
Ucf101 & 0 & 0.012 & 0.012 & 0.012 & 0.012 & 0.012 & 0.012 & 0.012 & 0.012 & 0.012 & 0.012 & 0.012 & 0.012 & 0.012 & 0.012 \\ \hline
Ucf101 & 1 & 1.322 & 1.200 & 1.033 & 1.333 & 1.367 & 1.300 & 1.350 & 1.033 & 1.120 & 1.367 & 1.567 & 1.167 & 1.600 & \textbf{6.000} \\ \hline
Ucf101 & 2 & 1.419 & 1.200 & 0.800 & 1.300 & 1.133 & 1.067 & 1.337 & 0.967 & 1.307 & 1.633 & 1.500 & 1.433 & 2.067 & \textbf{10.067} \\ \hline
Ucf101 & 4 & 1.357 & 1.167 & 0.967 & 1.500 & 1.433 & 1.233 & 1.480 & 1.000 & 1.727 & 1.333 & 1.500 & 1.367 & 3.100 & \textbf{13.033} \\ \hline
Ucf101 & 8 & 1.956 & 1.100 & 1.100 & 1.400 & 1.067 & 1.800 & 1.797 & 1.167 & 1.613 & 1.833 & 1.967 & 1.533 & 3.800 & \textbf{17.367} \\ \hline
Ucf101 & 16 & 2.696 & 1.233 & 1.333 & 1.900 & 1.533 & 1.667 & 1.667 & 1.667 & 2.097 & 1.533 & 2.200 & 1.800 & 7.433 & \textbf{22.800} \\ \hline
Cub & 0 & 0.005 & 0.005 & 0.005 & 0.005 & 0.005 & 0.005 & 0.005 & 0.005 & 0.005 & 0.005 & 0.005 & 0.005 & 0.005 & 0.005 \\ \hline
Cub & 1 & 0.399 & 0.500 & 0.650 & 0.500 & 0.533 & 0.500 & 0.490 & 0.500 & 0.533 & 0.533 & 0.500 & 0.700 & \textbf{1.233} & 0.900 \\ \hline
Cub & 2 & 0.556 & 0.500 & 0.600 & 0.533 & 0.467 & 0.433 & 0.503 & 0.600 & 0.490 & 0.633 & 0.633 & 0.533 & \textbf{1.333} & 1.133 \\ \hline
Cub & 4 & 0.541 & 0.500 & 0.900 & 0.467 & 0.567 & 0.333 & 0.533 & 0.467 & 0.490 & 0.633 & 0.567 & 0.500 & 1.300 & \textbf{1.500} \\ \hline
Cub & 8 & 0.712 & 0.500 & 0.800 & 0.500 & 0.400 & 0.433 & 0.510 & 0.433 & 0.500 & 0.600 & 0.533 & 0.467 & 1.533 & \textbf{1.700} \\ \hline
Cub & 16 & 1.061 & 0.500 & 0.700 & 0.533 & 0.467 & 0.367 & 0.483 & 0.400 & 0.537 & 0.600 & 0.567 & 0.567 & \textbf{2.133} & 1.967 \\ \hline \hline
Average & 0 & 0.008 & 0.008 & 0.008 & 0.008 & 0.008 & 0.008 & 0.008 & 0.008 & 0.008 & 0.008 & 0.008 & 0.008 & 0.008 & 0.008 \\ \hline
Average & 1 & 1.775 & 0.814 & 0.979 & 1.314 & 1.024 & 1.248 & 0.862 & 0.900 & 1.117 & 1.495 & 1.590 & 1.162 & 1.867 & \textbf{4.633} \\ \hline
Average & 2 & 2.561 & 0.838 & 1.062 & 1.790 & 1.033 & 1.529 & 0.863 & 1.081 & 1.308 & 2.143 & 2.619 & 1.462 & 2.338 & \textbf{7.038} \\ \hline
Average & 4 & 3.539 & 0.886 & 1.138 & 2.524 & 1.219 & 2.171 & 1.103 & 1.114 & 2.149 & 2.695 & 3.490 & 1.986 & 3.129 & \textbf{10.114} \\ \hline
Average & 8 & 4.802 & 0.962 & 1.290 & 2.952 & 1.324 & 3.010 & 1.053 & 1.143 & 2.849 & 3.629 & 4.105 & 2.605 & 4.429 & \textbf{14.024} \\ \hline
Average & 16 & 6.213 & 1.071 & 1.457 & 3.586 & 1.657 & 3.938 & 1.370 & 1.495 & 3.522 & 4.290 & 4.781 & 3.357 & 7.238 & \textbf{18.764} \\ \hline
Overall Average (excl. zs) & - & 3.778 & 0.914 & 1.185 & 2.433 & 1.251 & 2.379 & 1.050 & 1.147 & 2.189 & 2.850 & 3.317 & 2.114 & 3.800 & \textbf{10.915} \\ \hline
\end{tabular}}
\caption{Few-shot performance results for unlearned CLIP with \textit{90\%} level of knowledge lost evaluated across multiple datasets, methods, and shot counts. For shot counts greater than zero, the top-performing results are indicated in \textbf{bold}.  Average results are shown at the end of the table. The table also includes averages for each shot count, with a final overall average calculated by excluding the zero-shot results.}
\label{tab:few_shot_results}
\end{table}

\twocolumn

\clearpage
\subsection{Default Runs Without Unlearning}

\label{app:clip_nounl}

In this section, we present full results across different methods, number of shots and datasets on standard CLIP without unlearning. Aggregated results are presented and discussed in Sec. 5.2 of the main paper. SEPRES remains competitive also in a normal setting, showing overall best average performance. 

\begin{table}[h!]
\onecolumn
\centering
\resizebox{1.\textwidth}{!}{\begin{tabular}{|c|c||c|c|c|c|c|c|c|c|c|c|c|c|c|c|}
\hline
\rowcolor{gray!20} \textbf{Dataset} & \textbf{Shots} & \textbf{CLIPLora} & \textbf{CLIPAdapter} & \textbf{CoCoOp} & \textbf{CoOp} & \textbf{CoPrompt} & \textbf{IVLP} & \textbf{KgCoOp} & \textbf{MaPLe} & \textbf{ProGrad} & \textbf{PromptSRC} & \textbf{SEP} & \textbf{TCP} & \textbf{TaskRes} & \textbf{SEPRES} \\ \hline
Stanforddogs & 0 & 59.117 & 59.117 & 59.117 & 59.117 & 59.117 & 59.117 & 59.117 & 59.117 & 59.117 & 59.117 & 59.117 & 59.117 & 59.117 & 59.117 \\ \hline
Stanforddogs & 1 & 63.732 & 60.200 & 32.400 & 59.433 & 63.633 & 62.333 & 62.800 & \textbf{63.833} & 61.275 & 63.333 & 61.600 & 59.533 & 61.033 & 61.933 \\ \hline
Stanforddogs & 2 & 65.335 & 60.833 & 63.467 & 63.100 & 64.767 & 65.167 & 63.837 & 64.500 & 64.215 & \textbf{66.833} & 64.900 & 62.867 & 61.400 & 65.200 \\ \hline
Stanforddogs & 4 & 67.818 & 61.567 & 65.200 & 66.233 & 67.500 & 69.700 & 65.830 & 67.700 & 67.350 & \textbf{70.733} & 69.067 & 66.900 & 62.667 & 69.267 \\ \hline
Stanforddogs & 8 & 72.101 & 62.700 & 66.533 & 70.067 & 70.400 & 73.333 & 67.880 & 69.567 & 68.860 & \textbf{74.033} & 72.800 & 70.433 & 63.733 & 72.550 \\ \hline
Stanforddogs & 16 & 75.924 & 65.200 & 68.367 & 74.300 & 72.633 & 76.200 & 69.627 & 72.833 & 73.010 & \textbf{77.333} & 76.867 & 74.367 & 65.367 & 77.067 \\ \hline
Stanfordcars & 0 & 65.514 & 65.514 & 65.514 & 65.514 & 65.514 & 65.514 & 65.514 & 65.514 & 65.514 & 65.514 & 65.514 & 65.514 & 65.514 & 65.514 \\ \hline
Stanfordcars & 1 & \textbf{70.360} & 66.500 & 66.667 & 67.567 & 64.500 & 68.550 & 67.250 & 66.800 & 68.260 & 69.767 & 68.133 & 68.567 & 67.467 & 69.567 \\ \hline
Stanfordcars & 2 & 73.863 & 66.900 & 67.633 & 70.333 & 66.167 & 72.100 & 68.450 & 68.500 & 70.730 & 72.933 & 73.533 & 72.800 & 69.067 & \textbf{74.200} \\ \hline
Stanfordcars & 4 & 77.109 & 67.433 & 69.133 & 74.500 & 67.300 & 75.000 & 68.863 & 70.100 & 73.490 & 77.200 & 78.133 & 74.950 & 69.767 & \textbf{78.300} \\ \hline
Stanfordcars & 8 & 82.146 & 68.600 & 70.800 & 78.800 & 70.400 & 78.800 & 70.697 & 71.433 & 77.225 & 81.067 & \textbf{83.167} & 80.100 & 72.000 & 83.100 \\ \hline
Stanfordcars & 16 & 86.150 & 70.600 & 71.967 & 82.033 & 73.550 & 82.367 & 73.327 & 74.067 & 78.030 & 84.267 & 86.233 & 83.400 & 75.867 & \textbf{86.267} \\ \hline
Oxfordflowers & 0 & 70.767 & 70.767 & 70.767 & 70.767 & 70.767 & 70.767 & 70.767 & 70.767 & 70.767 & 70.767 & 70.767 & 70.767 & 70.767 & 70.767 \\ \hline
Oxfordflowers & 1 & 84.179 & 71.233 & 71.233 & 80.200 & 74.933 & 83.333 & 79.453 & 76.033 & 80.010 & 85.667 & 87.600 & 85.033 & 73.267 & \textbf{87.667} \\ \hline
Oxfordflowers & 2 & 89.809 & 71.733 & 51.967 & 88.100 & 78.167 & 89.900 & 79.847 & 79.233 & 81.973 & 91.133 & 91.967 & 90.467 & 73.800 & \textbf{92.100} \\ \hline
Oxfordflowers & 4 & 93.707 & 73.133 & 80.567 & 91.700 & 85.033 & 93.233 & 87.063 & 85.633 & 90.527 & 94.000 & \textbf{94.067} & 93.200 & 75.333 & 94.033 \\ \hline
Oxfordflowers & 8 & \textbf{96.590} & 75.867 & 83.767 & 94.467 & 91.033 & 95.533 & 87.793 & 90.433 & 92.950 & 95.900 & 96.300 & 95.733 & 77.033 & 96.200 \\ \hline
Oxfordflowers & 16 & \textbf{97.970} & 83.967 & 86.933 & 96.800 & 94.267 & 97.133 & 92.083 & 94.100 & 94.563 & 97.767 & 97.433 & 97.067 & 82.067 & 97.467 \\ \hline
Caltech101 & 0 & 93.306 & 93.306 & 93.306 & 93.306 & 93.306 & 93.306 & 93.306 & 93.306 & 93.306 & 93.306 & 93.306 & 93.306 & 93.306 & 93.306 \\ \hline
Caltech101 & 1 & 94.009 & 93.533 & 94.450 & 92.367 & \textbf{94.867} & 92.900 & 94.240 & 93.233 & 93.413 & 92.933 & 93.367 & 94.033 & 92.833 & 93.500 \\ \hline
Caltech101 & 2 & 94.821 & 93.667 & 94.350 & 93.800 & \textbf{94.900} & 93.833 & 94.590 & 94.500 & 94.807 & 94.833 & 94.500 & 94.667 & 93.067 & 94.600 \\ \hline
Caltech101 & 4 & 95.213 & 94.000 & 95.100 & 94.367 & 95.267 & 95.000 & 94.633 & 94.867 & 94.807 & 95.400 & \textbf{95.533} & 95.233 & 93.533 & 95.500 \\ \hline
Caltech101 & 8 & 95.754 & 94.467 & 95.067 & 95.267 & 95.867 & 95.800 & 94.943 & 95.167 & 95.417 & 95.767 & \textbf{96.200} & 95.600 & 94.033 & \textbf{96.200} \\ \hline
Caltech101 & 16 & 95.997 & 94.600 & 95.267 & 95.533 & 96.033 & 96.133 & 94.970 & 95.400 & 95.600 & 96.233 & 96.400 & 95.700 & 94.700 & \textbf{96.467} \\ \hline
Fgvcaircraft & 0 & 24.752 & 24.752 & 24.752 & 24.752 & 24.752 & 24.752 & 24.752 & 24.752 & 24.752 & 24.752 & 24.752 & 24.752 & 24.752 & 24.752 \\ \hline
Fgvcaircraft & 1 & 30.033 & 26.533 & 13.600 & 21.867 & 22.400 & 21.567 & 28.060 & 26.967 & 27.440 & 28.367 & 30.133 & 28.833 & 26.667 & \textbf{31.733} \\ \hline
Fgvcaircraft & 2 & 31.513 & 27.300 & 22.967 & 25.867 & 28.800 & 28.400 & 28.320 & 28.733 & 30.520 & 31.800 & 31.967 & 30.500 & 26.667 & \textbf{33.100} \\ \hline
Fgvcaircraft & 4 & 37.544 & 28.067 & 19.367 & 32.367 & 29.900 & 36.567 & 32.530 & 26.200 & 33.820 & 38.100 & 37.733 & 35.300 & 28.133 & \textbf{38.333} \\ \hline
Fgvcaircraft & 8 & 44.894 & 28.700 & 25.067 & 38.067 & 33.200 & 40.167 & 33.740 & 33.067 & 37.210 & 42.800 & \textbf{45.067} & 40.100 & 29.633 & 44.633 \\ \hline
Fgvcaircraft & 16 & \textbf{56.706} & 30.600 & 28.700 & 42.433 & 35.967 & 46.567 & 34.610 & 35.500 & 39.790 & 50.000 & 51.833 & 44.133 & 32.600 & 52.067 \\ \hline
Ucf101 & 0 & 67.460 & 67.460 & 67.460 & 67.460 & 67.460 & 67.460 & 67.460 & 67.460 & 67.460 & 67.460 & 67.460 & 67.460 & 67.460 & 67.460 \\ \hline
Ucf101 & 1 & \textbf{76.227} & 68.200 & 72.433 & 70.467 & 71.833 & 73.567 & 73.990 & 71.667 & 73.313 & 73.900 & 75.133 & 74.200 & 66.600 & 75.267 \\ \hline
Ucf101 & 2 & 78.624 & 69.267 & 73.433 & 73.533 & 75.467 & 75.900 & 74.913 & 73.367 & 75.007 & 77.867 & 78.600 & 77.400 & 67.233 & \textbf{78.900} \\ \hline
Ucf101 & 4 & 80.360 & 71.600 & 74.567 & 77.067 & 78.633 & 79.767 & 77.400 & 76.500 & 77.733 & 81.067 & 81.033 & 80.033 & 69.200 & \textbf{81.167} \\ \hline
Ucf101 & 8 & 83.725 & 73.500 & 77.367 & 80.267 & 81.100 & 82.933 & 78.977 & 79.167 & 79.733 & 84.267 & \textbf{84.567} & 83.667 & 71.467 & 84.267 \\ \hline
Ucf101 & 16 & 86.245 & 76.267 & 77.933 & 82.800 & 83.233 & 85.533 & 80.457 & 81.333 & 82.160 & 86.367 & \textbf{87.167} & 85.367 & 74.667 & 87.000 \\ \hline
Cub & 0 & 55.009 & 55.009 & 55.009 & 55.009 & 55.009 & 55.009 & 55.009 & 55.009 & 55.009 & 55.009 & 55.009 & 55.009 & 55.009 & 55.009 \\ \hline
Cub & 1 & \textbf{59.729} & 56.167 & 57.350 & 55.867 & 58.550 & 58.533 & 57.023 & 58.667 & 58.195 & 58.967 & 58.933 & 58.700 & 50.833 & 59.533 \\ \hline
Cub & 2 & 64.152 & 56.900 & 58.150 & 60.033 & 59.400 & 62.933 & 58.640 & 60.300 & 60.120 & 63.800 & 64.067 & 64.133 & 52.100 & \textbf{64.600} \\ \hline
Cub & 4 & 69.217 & 57.833 & 60.000 & 65.467 & 61.300 & 67.533 & 60.307 & 61.500 & 63.427 & 68.633 & \textbf{70.733} & 68.800 & 53.100 & 70.700 \\ \hline
Cub & 8 & 74.402 & 59.433 & 60.450 & 71.167 & 64.750 & 71.900 & 61.957 & 63.267 & 66.847 & 73.500 & 75.800 & 74.500 & 56.500 & \textbf{76.000} \\ \hline
Cub & 16 & 78.625 & 61.700 & 62.250 & 75.333 & 67.600 & 76.467 & 62.963 & 66.867 & 71.755 & 78.100 & 79.933 & 77.600 & 61.067 & \textbf{80.067} \\ \hline \hline
Average & 0 & 62.275 & 62.275 & 62.275 & 62.275 & 62.275 & 62.275 & 62.275 & 62.275 & 62.275 & 62.275 & 62.275 & 62.275 & 62.275 & 62.275 \\ \hline
Average & 1 & 68.324 & 63.195 & 58.305 & 63.967 & 64.388 & 65.826 & 66.117 & 65.314 & 65.987 & 67.562 & 67.843 & 66.986 & 62.671 & \textbf{68.457} \\ \hline
Average & 2 & 71.160 & 63.800 & 61.710 & 67.824 & 66.810 & 69.748 & 66.942 & 67.019 & 68.196 & 71.314 & 71.362 & 70.405 & 63.333 & \textbf{71.814} \\ \hline
Average & 4 & 74.424 & 64.805 & 66.276 & 71.671 & 69.276 & 73.829 & 69.518 & 68.929 & 71.593 & 75.019 & 75.186 & 73.488 & 64.533 & \textbf{75.329} \\ \hline
Average & 8 & 78.516 & 66.181 & 68.436 & 75.443 & 72.393 & 76.924 & 70.855 & 71.729 & 74.035 & 78.190 & \textbf{79.129} & 77.162 & 66.343 & 78.993 \\ \hline
Average & 16 & \textbf{82.517} & 68.990 & 70.202 & 78.462 & 74.755 & 80.057 & 72.577 & 74.300 & 76.415 & 81.438 & 82.267 & 79.662 & 69.476 & 82.343 \\ \hline
Overall Average (excl. zs) & - & 74.988 & 65.394 & 64.986 & 71.473 & 69.524 & 73.277 & 69.202 & 69.458 & 71.245 & 74.705 & 75.157 & 73.540 & 65.271 & \textbf{75.387} \\ \hline
\end{tabular}}
\caption{Few-shot performance results \textit{without} unlearning evaluated across multiple datasets, methods, and shot counts. For shot counts greater than zero, the top-performing results are indicated in \textbf{bold}.  Average results are shown at the end of the table. The table also includes averages for each shot count, with a final overall average calculated by excluding the zero-shot results.}
\label{tab:few_shot_results}
\end{table}

\twocolumn

\clearpage
\subsection{Aggressive Unlearning - Aggregated Results}
\label{app:aggunl_aggregated}

In Tab. \ref{tab:loss_values} we show the loss in accuracy relative to the default case (from Tab. \ref{tab:default_few_shot_results}) as we unlearn more knowledge from CLIP. When knowledge lost is 25\% SEPRES method loses less accuracy relative to its default performance compared to other methods. Similarly, when knowledge lost is 50\% SEPRES is still the best. For 90\% of lost knowledge all methods perform badly in few-shot setting. Note that we ignored TaskRes from calculating the minimum because it's default accuracy was lower compared to when a certain percentage of original knowledge was lost.

\begin{table}[htbp]
    \centering
    \onecolumn
    \resizebox{1.\textwidth}{!}{\begin{tabular}{|c|c|c|c|c|c|c|c|c|c|c|c|c|c|c|c|}
        \hline
       \rowcolor{lightgray} \textbf{Shots} & \textbf{Knowledge Lost (\%)} & CLIPLora & CLIPAdapter & CoCoOp & CoOp & CoPrompt & IVLP & KgCoOp & MaPLe & ProGrad & PromptSRC & SEP & TCP & TaskRes & SEPRES \\
        \hline
         1 & 25 & 46.393 & 39.693 & 44.010 & 48.275 & 45.915 & 47.532 & 38.538 & 31.646 & 36.275 & 49.128 & 48.560 & 50.219 & -35.583 & \textbf{24.143}\\
        \hline
        2 & 25 & 49.136 & 41.184 & 38.278 & 49.781 & 52.745 & 56.291 & 38.905 & 42.048 & 38.727 & 51.723 & 48.054 & 52.999 & -45.821 & \textbf{19.367}\\
        \hline
        4 & 25 & 45.775 & 44.353 & 39.065 & 48.340 & 52.789 & 50.998 & 44.517 & 43.683 & 40.789 & 48.976 & 45.628 & 51.485 & -60.087 & \textbf{15.192}\\
        \hline
        8 & 25 & 41.075 & 48.354 & 45.475 & 47.240 & 56.457 & 48.999 & 41.634 & 44.152 & 42.128 & 45.688 & 43.001 & 49.155 & -51.073 & \textbf{10.139}\\
        \hline
        16 & 25 & 35.920 & 50.079 & 50.372 & 46.102 & 51.600 & 44.160 & 47.555 & 54.887 & 43.415 & 40.370 & 39.104 & 45.911 & -14.370 & \textbf{6.861}\\
        \hline
        \hline
        1 & 50 & 54.714 & 52.497 & 41.076 & 55.293 & 51.913 & 42.078 & 58.495 & 39.836 & \textbf{34.017} & 56.189 & 53.971 & 58.696 & 28.221 & 39.940\\
        \hline
        2 & 50 & 55.410 & 52.015 & 51.452 & 55.467 & 58.601 & 57.614 & 50.388 & 50.602 & 36.707 & 56.198 & 55.336 & 62.739 & 20.461 & \textbf{33.292}\\
        \hline
        4 & 50 & 53.552 & 54.000 & 48.664 & 55.222 & 62.658 & 56.016 & 58.687 & 59.056 & 42.703 & 55.772 & 52.137 & 62.972 & 12.075 & \textbf{27.071}\\
        \hline
        8 & 50 & 49.098 & 57.099 & 57.730 & 52.860 & 65.434 & 54.279 & 60.981 & 54.366 & 43.598 & 51.686 & 48.904 & 59.897 & -2.122 & \textbf{20.029}\\
        \hline
        16 & 50 & 41.382 & 58.188 & 61.887 & 53.389 & 60.107 & 49.230 & 56.637 & 59.464 & 44.901 & 48.379 & 47.895 & 56.255 & 11.205 & \textbf{14.895}\\
        \hline
        \hline
        1 & 90 & 77.628 & 78.105 & 74.878 & 83.722 & 77.766 & 77.316 & 81.181 & \textbf{71.854} & 76.919 & 86.644 & 86.255 & 87.426 & 59.918 & 85.814\\
        \hline
        2 & 90 & 80.686 & 77.834 & 76.867 & 85.050 & 82.351 & 86.515 & 81.229 & \textbf{72.651} & 78.023 & 86.158 & 84.714 & 89.881 & 52.834 & 84.619\\
        \hline
        4 & 90 & 81.412 & 78.118 & 80.050 & 83.932 & 85.646 & 86.004 & 81.980 & 82.192 & \textbf{75.949} & 86.358 & 84.096 & 89.507 & 52.495 & 82.610\\
        \hline
        8 & 90 & 81.057 & 79.218 & 84.624 & 84.512 & 90.528 & 85.617 & 86.516 & 86.820 & \textbf{77.985} & 85.397 & 84.510 & 88.294 & 55.651 & 79.583\\
        \hline
        16 & 90 & 79.572 & 82.114 & 86.998 & 83.694 & 90.720 & 84.837 & 87.422 & 89.735 & 79.870 & 85.319 & 84.226 & 86.545 & 65.384 & \textbf{75.590}\\
        \hline
    \end{tabular}}
    \caption{Accuracy lost relative to the default case as a certain level of knowledge from CLIP is lost. SEPRES loses less accuracy relative to its default performance compared to other methods in most cases. }
    \label{tab:loss_values}
\end{table}

\end{document}